\documentclass{article}

\usepackage{microtype}
\usepackage{graphicx}
\usepackage{subfigure}
\usepackage{booktabs} 
\usepackage{tikz} 
\usepackage{multirow}

\usepackage{hyperref}

\usepackage[accepted]{icml2023}

\usepackage{amsmath}
\DeclareMathOperator{\Tr}{Tr}
\usepackage{amssymb}
\usepackage{mathtools}
\usepackage{amsthm}

\usepackage[capitalize,noabbrev]{cleveref}

\theoremstyle{plain}

\theoremstyle{definition}

\theoremstyle{remark}

\usepackage[textsize=tiny]{todonotes}

\icmltitlerunning{Improving Multimodal VAEs with Normalizing Flows and Correlation Analysis}

\begin{document}

\onecolumn
\icmltitle{Improving Multimodal Joint Variational Autoencoders
through Normalizing Flows and Correlation Analysis}



\icmlsetsymbol{equal}{*}

\begin{icmlauthorlist}
\icmlauthor{Agathe Senellart}{equal,aff}
\icmlauthor{Clément Chadebec}{aff}
\icmlauthor{Stéphanie Allassonnière}{aff}

\end{icmlauthorlist}

\icmlaffiliation{aff}{Université Paris Cité, INRIA, Inserm, Sorbonne Université, Centre de Recherche des Cordeliers }

\icmlcorrespondingauthor{Firstname1 Lastname1}{first1.last1@xxx.edu}
\icmlcorrespondingauthor{Firstname2 Lastname2}{first2.last2@www.uk}


\vskip 0.3in



\printAffiliationsAndNotice{}  

\begin{abstract}
We propose a new multimodal variational autoencoder that enables to generate from the joint distribution and conditionally to any number of complex modalities. 
The unimodal posteriors are conditioned on the Deep Canonical Correlation Analysis embeddings which preserve the shared information across modalities leading to more coherent cross-modal generations.
Furthermore, we use Normalizing Flows to enrich the unimodal posteriors and achieve more diverse data generation. Finally, we propose to use a Product of Experts for inferring one modality from several others which makes the model scalable to any number of modalities. We demonstrate that our method improves likelihood estimates, diversity of the generations and in particular coherence metrics in the conditional generations on several datasets.
\end{abstract}

\section{Introduction}

In many cases, information is conveyed through multiple heterogeneous modalities. 
For instance in the medical field, a patient status is described through
the results of several analyses: sonograms, MRI, blood analysis, etc... 
Two important challenges in multimodal machine learning are the task of \textit{learning relevant joint representations} and the task of \textit{generating realistic data}, either from one modality to another or in all modalities jointly. Generated data can for instance be used for data augmentation (DA) to improve the 
training of deep learning models on small datasets \cite{tanner1987calculation,shorten2019survey,chadebec_data_2022}. For instance, augmenting jointly multiple modalities can be useful to improve the performance of deep segmentation networks \cite{li2020tumorgan}. In certain contexts,  conditional generation has also proved more relevant than unconditional generation to augment a specific modality since it allows benefiting from potentially useful information contained in another modality \cite{wei2019generative}. 
Data generation has also been used to anonymize a dataset \cite{shin2018medical} by replacing true individuals with synthetic ones with the same characteristics. Mathematically speaking we aim at \textit{modelling and sampling from the joint and conditional distributions} for any number of modalities of complex data. Good modelling implies preserving the shared information across modalities (referred to as \textit{coherence}) while keeping the diversity of each domain. 

Deep Generative Models, especially Generative Adversarial Networks (GAN) \cite{goodfellow_generative_2014} and Variational AutoEncoders (VAEs) \cite{kingma2013auto} are powerful tools for learning distributions of complex data such as images. In recent years, many multimodal variants of those models were developed. On the one hand, Multimodal GANs such as BicyleGAN \cite{zhu2017toward} or Pix2Pix \cite{isola2017image} perform well in the image translation setting where we aim at finding a mapping between different domains. However, there is no explicit statistical model of the underlying distributions leading to poor interpretability and their adversarial structure makes them hard to train \cite{saxena2021generative}. Furthermore, those models cannot be used to generate all modalities jointly. On the other hand, VAE based models rely on latent variables and are trained to optimize the model's likelihood with Variational Inference. There exist several approaches proposing to extend the VAE framework to multimodal datasets \cite{suzuki2022survey}. First, some works proposed to use coordinated unimodal VAEs and force the latent spaces for each modality to be similar \cite{wang2016deep,yin2017associate}. With such models, we can infer the latent variables from each modality individually but not from all of them simultaneously. However, it is desirable for the latent representation to be enriched by information coming from all modalities. To this end, joint models using a single latent space to explicitly extract a latent representation common to all modalities have been proposed. Among those models, one popular approach is to aggregate the unimodal inference distributions relying on a Mixture of Experts \cite{shi2019variational} or Product of Experts \cite{wu2018multimodal, shi2019variational,sutter_generalized_2021}. However, such aggregation may restrict the diversity of the generations \cite{daunhawer_limitations_2022}. Furthermore, those models can suffer from modality
collapse during training which requires careful gradient rescaling \cite{javaloy2022mitigating}. Finally, another approach is to have a dedicated encoder for the joint inference distribution. For instance the JMVAE\cite{suzuki2016joint} or the TELBO \cite{vedantamgenerative} models
do not suffer from lack of diversity and are less prone to modality collapse but 
have lower coherence than aggregated models \cite{shi2019variational} and are considered not easily scalable\cite{wu2018multimodal}.

In this article, we propose and justify a new VAE-based model that can be used to model the \textbf{joint, marginal and conditional distributions} of any number of modalities. This model proves to be able to produce more diverse and relevant samples than existing methods on several benchmark datasets, especially for the conditional generation task. In particular, our contributions are as follows:
\begin{itemize}
    \item We propose to use Normalizing Flows \cite{rezende2015variational} to model and improve the expressiveness of the unimodal posterior distributions used for cross-modal data generation. 
    \item We introduce a variant of this model relying on Deep Canonical Correlation Analysis \cite{andrew2013deep} to extract the shared information between modalities and propose to use posterior distributions conditioned on the DCCA embeddings instead of the data itself.
    \item We discuss and empirically show that these models are scalable to any number of modalities by using a Product of Experts to model distributions conditioned on a subset of modalities. 
    \item We extensively test our methods on several benchmark datasets and show that the proposed models outperform several \textit{state-of-the-art} methods in terms of cross-modal data generation coherence and diversity as well as likelihood estimates. 
\end{itemize}

\section{Background}
In this section, we recall mathematical tools relevant to this paper, starting with the Joint Multimodal Variational Autoencoder framework. 

\subsection{Joint Multimodal Variational Autoencoders}
The Joint Multimodal Variational Autoencoder (JMVAE) \cite{suzuki2016joint} model is one of the first extension of the VAE to the multimodal setting. Let $\mathcal{X} = \{ X^{(i)} \}_{i=1}^N$ be a set of $N$ i.i.d observations where each observation comprises $m$ modalities $X^{(i)} = (x_1^{(i)}, x_2^{(i)}, ..., x_m^{(i)})$. We aim at modelling the joint distribution $p(X)$, and for any $i \in [1,m]$ and $S$, a subset of $[1,m]\backslash\{i\}$ the conditional distribution $p(x_i|(x_s)_{s \in S})$.
For readability, in the following we only consider two modalities \textit{i.e }$m=2$.
In the VAE setting, we consider that each observation $X=(x_1,x_2)$ is sampled from  the following latent generative model :
\begin{equation}
\begin{split}
    p_{\theta}(x_1,x_2, z) &= p_{\theta}(x_1,x_2|z)p(z)\\
     &= p_{\theta}(x_1|z)p_{\theta}(x_2|z)p(z)\,,
\end{split}
\end{equation}
where we assume the modalities to be independent when conditioned on $z$, $p(z)$ is a prior distribution over the latent variables and $p_{\theta}(x_i|z)$ with $i\in \{1, 2\}$ are the unimodal generative distributions often referred to as the \emph{decoders}. $p(z)$ is usually chosen as a standard normal distribution $\mathcal{N}(0,I)$ and $p_\theta(x_1|z),p_\theta(x_2|z)$ are chosen depending on the input data (\emph{e.g.} Bernoulli for binary data) and parameterized by deep neural networks. The first objective of the JMVAE model is to find a set of parameters $\theta \in \Theta$ that maximizes the likelihood of the observations $p_{\theta}(x_1, x_2)$.
Since that objective is often intractable, one can rely on variational inference  \cite{jordan1999introduction} and introduce a parametric distribution $q_\phi(z|x_1,x_2)$ referred to as the \emph{joint encoder} aiming at approximating the true posterior $p_\theta(z|x_1,x_2)$. That allows to derive a lower bound (ELBO):
\begin{equation}
\label{elbo def}
    \begin{split}
        \ln p_{\theta}(x_1,x_2) & \geq \ln \mathbb{E}_{q_{\phi}(z|x_1, x_2)} \left [ \frac{p_{\theta}(x_1,x_2|z)p(z)}{q_{\phi}(z|x_1,x_2)}\right ]\\
        &\geq \mathbb{E}_{q_{\phi}(z|x_1,x_2)}\left[\ln \frac{ p_{\theta}(x_1,x_2|z)}{q_\phi(z|x_1,x_2)} \right] \\
        &= \mathcal{L}(x_1,x_2)\,.
    \end{split}
\end{equation}
This bound can be optimized using the stochastic gradient descent algorithm.
For cross-modal data generation, we also want to model the conditional distributions $p_\theta(x_i|x_j)$ where $i, j \in\{1, 2\}$ and $i\neq j$ so that we can generate one modality from the other. To do so, one can write :
\begin{equation}
    \label{decomposition conditional dist}
    \begin{aligned}
    p_\theta(x_i, z|x_j) &=  p_\theta(x_i|z,x_j) p_\theta(z|x_j)\\
    &=p_\theta(x_i|z)p_\theta(z|x_j)\,.
    \end{aligned}
\end{equation}
Since the true unimodal posterior $p_\theta(z|x_j)$ is unknown, an auxiliary distribution $q_{\phi_j}(z|x_j)$ (called the \emph{unimodal encoders}) can be used to approximate it. Once the unimodal encoders are trained, we will be able to generate $x_i$ from $x_j$ by sampling first $z \sim q_{\phi_j}(z|x_j)$ and then $x_i \sim p_\theta(x_i|z)$. In the JMVAE model, auxiliary distributions are optimized using the following objective :
\begin{equation} 
    \label{ljm}
    \mathcal{L}_{\mathrm{JM}}(x_1,x_2) = \sum_{i \in \{1,2\}} \mathrm{KL}(q_{\phi}(z|x_1,x_2) || q_{\phi_i}(z|x_i)) \,.
\end{equation}
Finally, a combination of $\mathcal{L}_{JM}$ and the ELBO is optimized:
\begin{equation}
    \mathcal{L}_{\mathrm{JMVAE}}(x_1,x_2) = \mathcal{L}(x_1,x_2) - \alpha \mathcal{L}_{\mathrm{JM}}(x_1,x_2) \,,
\end{equation}
where $\alpha$ is a hyperparameter that regularizes the importance of the $\mathcal{L}_{\mathrm{JM}}$ term. \cite{suzuki2016joint} mentioned that $\alpha$ controls a trade-off between the quality of the reconstructions and the conditional generations. A high value of $\alpha$ imposes a strong regularization on $q_\phi$ and undermines the reconstruction term of the ELBO while a low value of $\alpha$ causes the unimodal encoders to be undertrained. 
In practice, the unimodal encoders are chosen as multivariate Gaussian distributions which may be too restrictive to model the true posteriors of complex data. In this paper, we propose to increase the expressiveness power of those distributions by using Normalizing Flows. Moreover, as shown in Eq.~\eqref{decomposition conditional dist}, in this model, estimating the conditional distributions relies on learning the posterior distributions conditioned directly on the data $x_i$. However, inferring one modality from another may not require the entire data but only the relevant information shared by the modalities. Hence, we propose to extract that information using Deep Canonical Correlation Analysis (DCCA) and to consider posterior distributions conditioned on the DCCA embeddings.

\subsection{Normalizing Flows}

Normalizing Flows are a powerful modelling tool that allows to model complex, differentiable distributions. They have been introduced by \cite{rezende2015variational} to improve variational inference in the classical VAE scheme. A flow is an invertible smooth transformation $f$ that can be applied to an initial distribution to create a new one, such that if $Z$ is a random vector with density $q(z)$, then $Z' = f(Z)$ has a density given by:
\begin{equation}
    q'(z') = q(z) \left|\det \frac{\partial f^{-1}}{\partial z'}\right| = q(z) \left |\det \frac{\partial f}{\partial z}\right |^{-1}\,.
\end{equation}
Combining $K$ transformations $z_K = f_K \circ f_{K-1} \circ ... \circ f_1(z_0)$ allows to gain in complexity of the final distribution.

\subsection{Deep Canonical Correlation Analysis}

Deep Canonical Correlation Analysis \cite{andrew2013deep} (DCCA) aims at finding correlated neural representations for two complex modalities such as images. 
It is based upon the classical Canonical Correlation Analysis (CCA) which we briefly recall here. Let $(X_1, X_2) \in \mathbb{R}^{n_1} \times \mathbb{R}^{n_2}$ two random vectors, $\Sigma_{1}, \Sigma_{2}$ their covariances matrices and $\Sigma_{1,2} = \mathrm{Cov}(X_1,X_2)$. CCA's objective is to find projections  $a^TX_1$, $b^TX_2$ that are maximally correlated :
\begin{equation*}
(a^*, b^*)  = \underset{a^T\Sigma_{1}a = b^T\Sigma_{2}b= 1}{\arg\max} a^T \Sigma_{1,2}b \,.
\end{equation*}
Once these optimal projections are found, we can set $(a_1,b_1) = (a^*, b^*)$ and search for subsequent projections $(a_i, b_i)_{2 \leq i \leq k}$ with the additional constraint that they must be uncorrelated with the previous ones. 
We can rewrite the problem of finding the first $k$ optimal pairs of projection as finding matrices $A \in \mathbb{R}^{(n_1,k)}$, $B \in \mathbb{R}^{(n_2,k)}$ that solve:
\begin{equation}
    (A^*, B^*)  = \underset{A^T\Sigma_{1}A = B^T\Sigma_{2}B = I}{\arg\max} \Tr (A^T \Sigma_{1,2}B)
\end{equation}
If we further have $k=n_1=n_2$ then the optimal value is $F(X_1,X_2) =\Tr(T^{\top}T)^{\frac{1}{2}}$ with $T\coloneqq \Sigma_1^{\frac{1}{2}}\Sigma_{1,2} \Sigma_{2}^{\frac{1}{2}}$. This value is the total CCA correlation of the random vectors $X_1,X_2$. It can also be seen as the sum of the singular values of $T$, each singular value representing the correlation of the embeddings along a direction. Note that $F$ only depends on the covariance matrices $(\Sigma_1, \Sigma_2, \Sigma_{1,2})$. With the DCCA we consider two neural networks $g_1$, $g_2$ so as to optimize the total CCA correlation $F(g_1(X_1), g_2(X_2))$. The gradient of this objective with respect to the parameters of $g_1, g_2$ can be computed so that gradient descent can be used.

When considering more than two modalities, a proposed extension to the CCA is to optimize the sum of the pairwise CCA objectives \cite{kanatsoulis2018structured}. We can adapt this idea to the DCCA framework and train DCCA encoders for $m$ modalities by maximizing $\sum_{i<j \in [|1,m|]} F(g_i(X_i),g_j(X_j))$.

\section{Proposed Model}
In this section, we introduce a new generative model for multimodal data inspired by the JMVAE framework but we propose to enhance the unimodal conditional distributions using normalizing flows. We also propose a variant in which these distributions are only conditioned using the learned DCCA embeddings of the modalities. Finally, in the case where we want to perform generation conditioned on more than one modality, we propose to use a Product of Experts of the unimodal posteriors.

\subsection{Learning the Joint Distribution}
Let $\mathcal{X} = \{ X^{(i)} \}_{i=1}^N$ be a set of observations composed of $m$ modalities $X^{(i)} = (x_1^{(i)},  \dots, x_m^{(i)})$. Keeping the same notations as before, we denote the joint encoder as $q_{\phi}(z|X)$ and decoder of modality $i$ as $p_{\theta}(x_i|z)$. The proposed model relies on a two steps training. First we optimize the ELBO according to Eq.~\eqref{elbo def} to learn the joint posterior distribution and the decoder distributions. In a second step, we learn the unimodal posterior distributions $(q_{\phi_i}(z|x_i))_{i\in[|1,m|]}$ by optimizing Eq.~\eqref{ljm} while keeping the joint encoder and decoder models (\textit{i.e} $\theta, \phi$) fixed. Contrary to the JMVAE method, we decided to rely on a two-step training since it allows to mitigate the trade-off induced by the parameter $\alpha$, the value of which can strongly influence the model's performances \cite{suzuki2016joint}. Furthermore, we empirically observed that a two-steps training did not hinder the performance of the JMVAE model and often improved it. This is discussed in Appendix \ref{training_compare}.
We assume that we have learned the joint encoder as well as the decoders and now detail how we propose to learn the unimodal posterior distributions needed to perform cross-modality data generation. 



\subsection{Integrating Normalizing Flows}
In order to properly estimate the true unimodal posterior distributions $(p_{\theta}(z|x_i))_{i\in[|1,m|]}$, the unimodal posteriors $(q_{\phi_i}(z|x_i))_{i\in[|1,m|]}$ need a lot of flexibility. However, those distributions are often chosen as multivariate Gaussian which may be a too restrictive class of distributions and may cause incoherence in the conditional distributions. This phenomenon is illustrated later on a toy dataset in Sec.~\ref{toy dataset section}. In our model, we propose to address this limitation by enriching these distributions using Normalizing Flows. More explicitly the expression of the unimodal distribution writes:
\begin{equation}
\label{flots mvae}
    \begin{split}
    \ln q_{\phi_i}(z_K|x_i) = \ln q_{\phi_i}^{(0)}(z_0|x_i) - \sum_{k = 1}^{K} \ln \left| \det \frac{\partial f_k^{(i)}}{\partial z_{k-1}}\right|\,,
    \end{split}
\end{equation}
where $q_{\phi_i}^{(0)}(z_0|x_i)$ is a simple parametrized distribution, the parameters of which are given by neural networks and $(f_k^{(i)})$ are Normalizing Flows. In practice, we use multivariate Normal distributions with diagonal covariance for $(q_{\phi_i}^{(0)})_{i\in[|1,m|]}$. We use Eq.~\eqref{ljm} to train the unimodal encoders. Since $q_{\phi}(z|X)$ is fixed at this point of the training this objective can be rewritten as :
\begin{equation}
\label{rewrite_ljm}
    \mathcal{L}_{\mathrm{JM}}(X) 
    = -\sum_{i = 1}^{m} \mathbb{E}_{q_\phi(z|X)}\left( \ln q_{\phi_i}(z|x_i) \right)\,.
\end{equation}
For $i\in[|1,m|]$, the expectation inside the sum can be estimated using samples from the joint encoder $q_\phi(z|X)$ and evaluating the density $\ln q_{\phi_i}(z|x_i)$ for those samples. Since we only need to perform density evaluation for $q_{\phi_i}(z|x_i)$ during the training, we choose to use Masked Autoregressive Flows (MAF) \cite{papamakarios2017masked} that allow to compute it efficiently. Eq.~\eqref{rewrite_ljm} shows that during the training, the unimodal encoders are \textit{informed} by the joint encoder: for each sample $X=(x_1,\dots,x_m)$, a latent variable $z$ is sampled from $q_{\phi}(z|X)$ and then for each $i\in[|1,m|]$, the probability $q_{\phi_i}(z|x_i)$ is maximized. Interestingly enough, by integrating Eq.~\eqref{rewrite_ljm} on the entire training set, we can show that $q_{\phi_i}(z|x_i)$ is encouraged to be close to an average distribution $q_{\mathrm{avg}}(z|x_i) =  \mathbb{E}_{\hat{p}((x_j)_{j\neq i}|x_i)}(q_\phi(z|X))$ where $\hat{p}$ is the observed empirical distribution of the data. 
This provides an intuition of the interest of $L_{JM}$ to fit the unimodal encoders.
This interpretation is detailed in Appendix.~\ref{interpretations}.

\subsection{Conditioning on the DCCA Embeddings}
\begin{figure}[h!]
    \centering
    \includegraphics[width=0.5\linewidth]{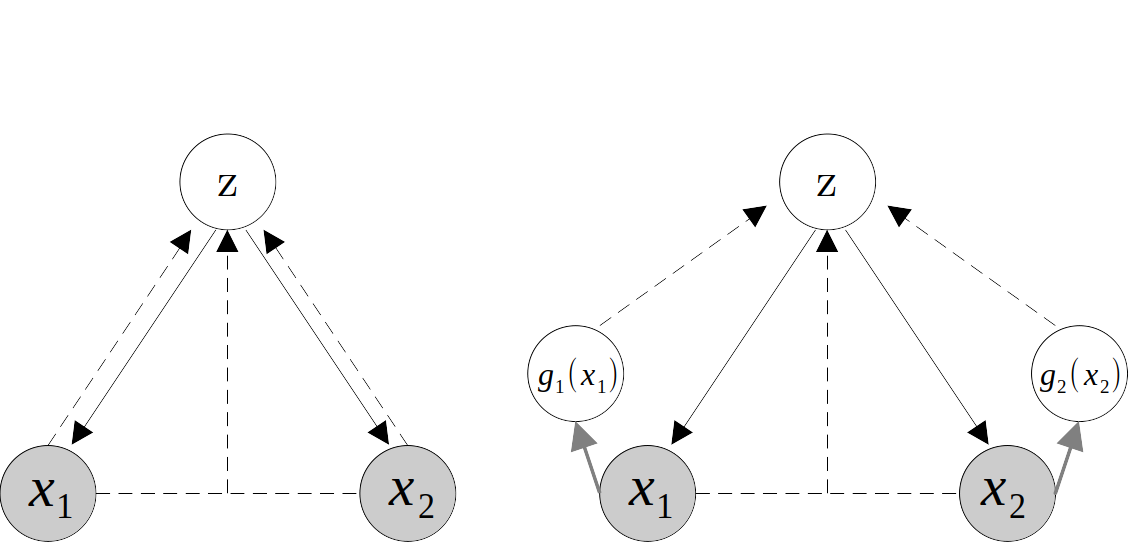}
    \caption{\emph{Left}: Graphical model for the JMVAE and JNF. \emph{Right}: Graphical model for the JNF-DCCA. The dashed lines represent encoders while the plain ones represent decoders. The grey bold arrows represent the pre-trained deterministic DCCA encoders.}
    \label{graph summary}
\end{figure}

A second observation is that to generate a modality from another one we only need the information shared by both and not the entire data. For instance, let us assume that we have two modalities of data representing a digit in different ways (\emph{e.g.} MNIST \cite{lecun_gradient-based_1998}-SVHN \cite{netzer_reading_2011}), we would only need the label of the digit to be able to generate from one domain to the other. In most cases, the information shared by the modalities is unknown but there are methods that aim to extract it \cite{hardoon_canonical_2004,tian2020contrastive}. 
Assume that for $i \neq j \in[|1,m|]$ we have a function $g_j$ such that $p_\theta(x_i, z|x_j)=p_\theta(x_i, z|(g_j(x_j)))$. Morally speaking, $g_j$ extracts the shared information between modalities while tuning out the modality specific information. Then, Eq.~\eqref{decomposition conditional dist} rewrites:
\begin{equation}
\label{DCCA decomposition}
    \begin{split}
        p_\theta(x_i, z|x_j)&=p_\theta(z|g_j(x_j)) p_\theta(x_i|z, g_j(x_j))\\
        &=p_\theta(z|g_j(x_j)) p_\theta(x_i|z)\\
    \end{split}
\end{equation}
Therefore, by making use of such functions $(g_j)_{j \in [|1,m|]}$, we can estimate the posteriors $p_\theta(z|g_j(x_j))$
that are eventually easier to model provided that we choose relevant functions. 
Building on that intuition, in this paper we choose to use DCCA representations as such functions as it is a versatile method that can be applied to various datasets.  
In the following, we refer to the model using only flows as JNF and to the model using both flows and DCCA embeddings as JNF-DCCA. Noteworthy is the fact that using the DCCA does not increase significantly the number of parameters since it reduces the dimension of the data reducing the number of parameters of the unimodal encoders $q_{\phi_j}(z|g_j(x_j))$. Graphical models are provided in Figure \ref{graph summary} for both methods. In JNF-DCCA, two representations of the data are actually extracted: the joint encoder provides latent variables $z$ containing all the information for both modalities while the DCCA only accounts for the correlated variables across modalities. 
\begin{algorithm}[h]
   \caption{Training of JNF, JNF-DCCA}
   \label{alg:training}
\begin{algorithmic}[1]
   \STATE {\bfseries Input:} Multimodal dataset with $m$ modalities $(x_1, x_2, ..,x_m)$
   \STATE Train the joint encoder $q_{\phi}(z|X)$ and decoders by maximizing the joint ELBO (Eq.~\eqref{elbo def}).
    \STATE If using DCCA : train the DCCA encoders $(g_i)_{i \in [|1,m|]}$ 
   \STATE Freeze $\phi, \theta$ and train the unimodal encoders by minimizing $\mathcal{L}_{\mathrm{JM}}(X)$ Eq.~\eqref{rewrite_ljm}.

\end{algorithmic}
\end{algorithm}

\subsection{Extending the Methodology to More Modalities}

It has been argued \cite{wu2018multimodal, shi2019variational} that the JMVAE is not a scalable model when more than two modalities are considered since for each subset $S$ of the $m$ modalities, approximating the posterior $p_\theta(z|(x_i)_{i \in S})$ would require introducing another encoder model. In this paper, we argue that we can actually use a Product of Experts (PoE) of well fitted unimodal posteriors to approximate it. Indeed, \cite{wu2018multimodal} show that: 
\begin{equation}
\label{why poe}
    \begin{split}
        p_\theta(z|(x_i)_{i \in S}) \propto \frac{\prod_{i \in S} p_\theta(z | x_i)}{p(z)^{|S|-1}}\,.
    \end{split}
\end{equation}
Assuming that the unimodal posteriors are such that $q_{\phi_i}(z|x_i) \approx p_\theta(z|x_i)$ we can use an approximation of $p_\theta(z|(x_i)_{i \in S})$ as follows:
\begin{equation}
\label{approw poe}
    \begin{split}
        q(z|(x_i)_{i \in S}) \propto \frac{\prod_{i \in S} q_{\phi_i}(z | x_i)}{p(z)^{|S|-1}}\,.
    \end{split}
\end{equation}
Using the DCCA, we can apply the same reasoning to model 
\begin{equation}
\label{approx poe}
    \begin{split}
        q(z|(g_i(x_i))_{i \in S}) \propto \frac{\prod_{i \in S} q_{\phi_i}(z | g_i(x_i))}{p(z)^{|S|-1}}\,.
    \end{split}
\end{equation}
In our method, we choose to model the unimodal posteriors $q_{\phi}(z|x_i)$ or $q_{\phi}(z|g_i(x_i))$ with Normalizing Flows. Therefore, the PoE in Eq.~\eqref{approw poe} does not have a closed form but we can easily sample from it using Hamiltonian Monte-Carlo Sampling \cite{neal2005hamiltonian}. A reminder of the HMC method and details on how it is used within our method is given in the Appendix.~\ref{app:hmc}. 
Note that this sampling is only needed at inference time and not during training. Therefore it does not impact the training time or complexity of the model. 



\section{Related Works}
 Several models were built on the same architecture as the JMVAE. The TELBO \cite{vedantamgenerative} model also uses a joint encoder
 and unimodal encoders for each modality but those are fit using the Triple ELBO Loss.
First the joint ELBO term is optimized then the decoders are fixed while the two unimodal ELBOs are optimized. The $M^2VAE$ \cite{korthals2019multi} takes from both the JMVAE and the TELBO and combines their losses.
These models encounter the same limitations as the JMVAE. First the unimodal posteriors are modelled as normal distributions which restrict the expressiveness of the cross-modal inference. The second limitation is their scalability, as no other solution were considered but to introduce a new network to model each subset posterior $p_\theta(z|(x_i)_{i \in S})$ for $S \in \mathcal{P}([|1,m|])$. In our method, we show that it is not necessary.
 
To solve this scalability issue, aggregated models \cite{suzuki2022survey} were suggested and model the joint posterior as a function of the unimodal posteriors. The first model to use that approach is the MVAE \cite{wu2018multimodal} model that builds upon Eq.\eqref{why poe} to suggest modelling the joint encoding distribution as a Product of Experts of the 
unimodal encoders and optimizing the ELBO (Eq.~\eqref{elbo def}).
In a similar fashion, the MMVAE model \cite{shi2019variational} uses a Mixture of Experts of the unimodal encoders and optimizes the Importance Weighted bound \cite{burda2015importance} instead of an ELBO. 
Finally the MoE-PoE  \cite{sutter_generalized_2021} combines the two approaches with a Mixture of Product of Experts. Aggregated models use fewer parameters but the resulting conditional distributions have been shown to be less diverse than the real ones due to the  constraints imposed on the unimodal encoders \cite{daunhawer_limitations_2022}. These models are also more sensitive to modality collapse as they integrate more \textit{impartiality blocks} causing conflictual gradients \cite{javaloy2022mitigating}. 

To palliate to these observations, recent models have used multiple latent spaces \cite{hsu2018disentangling,sutter2020multimodal,daunhawer2021self} to separate the shared and modality-specific information. However, those models are sensible to the \textit{shortcut} issue meaning that shared information leaks into the modality specific latent space leading to a poorer coherence \cite{palumbo2022mmvae+}.  

On the contrary, our approach only uses one latent space and few hyperparameters. 
As the lack of diversity comes from the aggregation, we use a dedicated encoder for the joint posterior. We train the unimodal encoders using flows and DCCA embeddings to improve coherence and diversity. For the conditional subsets posteriors, however, we follow the MVAE insight and use the PoE of the unimodal posteriors. This aggregation is only performed during inference and not during training to avoid modality collapse and lack of diversity in the unimodal posteriors. 

\section{Experiments}
In this section, we present the main results obtained on three datasets. Ablations studies are available in Appendices~\ref{app:dcca_embeddings} and \ref{app:number of flows}.
\subsection{A Toy Dataset}
\begin{figure*}[ht]
\begin{center}
\centerline{\includegraphics[width=\textwidth]{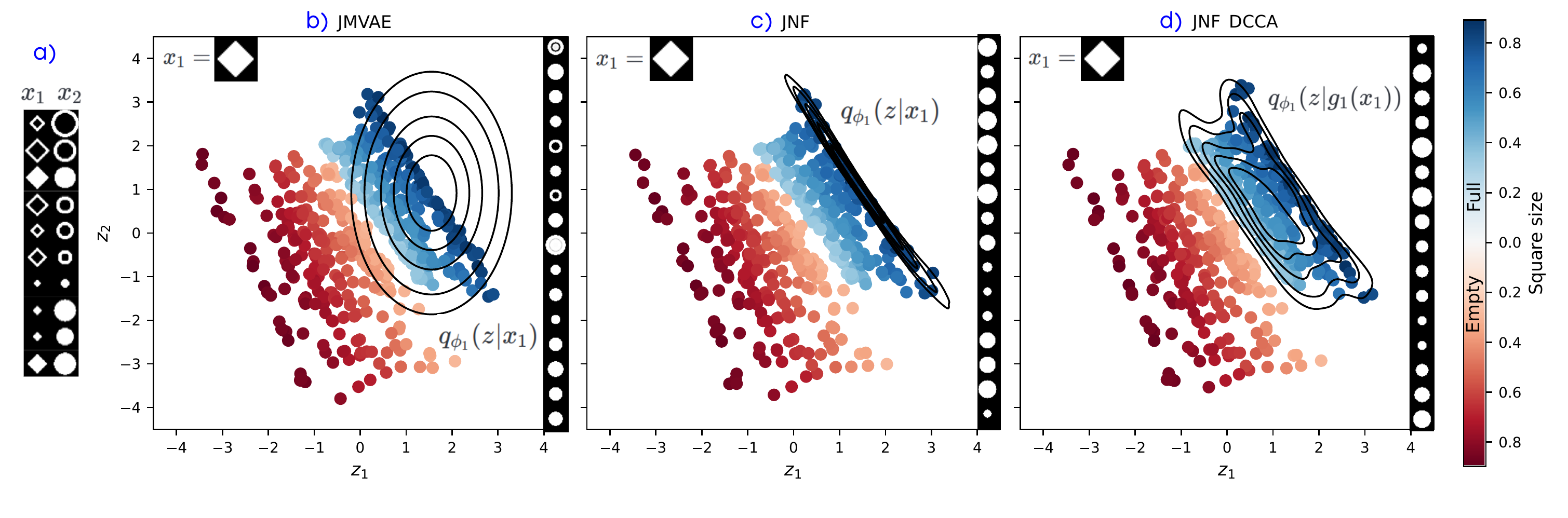}}
\caption{a) Samples from the toy dataset. b),c),d): For JMVAE, JNF and JNF-DCCA, we plot the joint embeddings of 100 training samples and the density of a single unimodal posterior. Each point correspond to a pair of images. The color (blue/red) of the point indicates if the images were empty or full, its intensity correspond to the size of the square in the pair. Altogether, we can see the distribution of the squares in the latent space given their classes (empty/full) and sizes. In particular, the deep blue points indicates the domain of the latent space that pertains to the square $x_1$ shown in the top-left corner.
For each model, we plot an example of $q_{\phi_1}(z|x_1)$. The JMVAE posterior encompasses the domain of the latent space that decodes into $x_1$ but also covers a large space that does not decode into $x_1$. That is because the Gaussian form is too restrictive. On the contrary, the JNF posterior covers only the right domain. On the right, we visualize the JNF-DCCA posterior $q_{\phi_1}(z|g_1(x_1))$. The DCCA of $x_1$ extracts the shared information between modalities \textit{i.e} the class. The resulting posterior covers the domain in the latent space that corresponds to the class of $x_1$ \textit{i.e} the blue points (full). On the right side of each plot, we show circles generated by sampling $z\sim q_{\phi_1}(z|x_1)$ (or $z\sim q_{\phi_1}(z|g_1(x_1))$) and decoding with $p_\theta(x_2|z)$.}
\label{toy_latent_space}
\end{center}
\vskip -0.2in
\end{figure*}
\label{toy dataset section}
First, we illustrate our contributions on a toy bimodal dataset of 32x32 images. The first modality is composed of images of squares while the second modality contains circles. The two modalities share the information of the shape being full or empty, but the sizes of each shape are independent. Figure~\ref{toy_latent_space}.a) presents a few samples from this dataset. 
We train the JMVAE on this dataset as well as our two proposed models, the JNF and JNF-DCCA. The simplicity of the dataset allows to use a 2-dimensional latent space. We use the same networks and two-steps training for all models. The first step is shared across the models, so to ease comparison, we train the joint encoder and the decoders once and use the same networks and joint latent space in all three models. The differences between the models lies in the modelling of the unimodal encoders. Figure \ref{toy_latent_space} presents a visualization of the latent space and an example of a learned unimodal posterior for each model. With this simple 2D example, we can see the latent domain that corresponds to a given image $x_1$ which aims to be appropriately covered by the unimodal encoders $q_{\phi_1}(z|x_1)$.
 In Figure \ref{toy_latent_space}, this domain (given by the dark blue points) cannot be approximated with a Gaussian, therefore the JMVAE fails to approximate well the true posterior. As a result, the conditional generation produces \textit{incoherent} results \textit{i.e} empty circles generated from full squares. The JNF posterior, enriched by normalizing flows, is more flexible and provides a better approximation.
We also illustrate the effect of using the DCCA. On the right of Figure \ref{toy_latent_space}, we see that $q_{\phi_1}(z|g_1(x_1))$ conditioned on the DCCA embeddings, covers all the domain that decodes into full pairs. When inferring an image $x_2$ from $x_1$, all we need to know is whether $x_1$ is full or empty since the sizes of the shapes are independent. Therefore $q_{\phi_1}(z|g_1(x_1))$ can be used to infer $x_2$ from $x_1$. In this simple example, both posteriors $q_{\phi_1}(z|x_1)$ and $q_{\phi_1}(z|g_1(x_1))$ are easy to approximate with Normalizing Flows. However, in certain cases $q_{\phi_1}(z|g_1(x_1))$ may have a simpler shape than $q_{\phi_1}(z|x_1)$ and so be easier to approximate. For instance, in the MNIST-SVHN dataset presented below, JNF-DCCA outperforms significantly other models.
\subsection{Benchmark Datasets}
We then test our methods on several benchmark datasets used in previous studies \cite{suzuki2016joint, wu2018multimodal,shi2019variational}.
First, we use the MNIST-SVHN dataset: each image is paired with 5 different images with the same label in the other modality. 
We also use CelebA-64 \cite{liu2015deep} in which we consider the images to be the first modality and the binary vector of 40 attributes as a second modality. 
Finally, we create a trimodal dataset by pairing MNIST, SVHN and FashionMNIST \cite{xiao2017fashion} to test the scalability of our method. We do 5 different random matching between images sharing the same label.

\subsection{Experiments Settings}
We compare our results to the JMVAE, MVAE and MMVAE. For fair comparison, the same architectures are used for the unimodal encoders $(q_{\phi_i}(z|x_i))_{1 \leq i \leq m}$ in all models but the JNF-DCCA in which the same encoders are used for the DCCA $(g_i)_{1 \leq i \leq m}$ and the $q_{\phi_i}(z|g_i(x_i))$ are simple MLPs. 
All the models architectures and training specifics are summarized in Appendix \ref{architectures} for reproducibility.
Note that in the original MMVAE formulation, the IWAE \cite{burda2015importance} bound is optimized which is known to produce better likelihoods than the ELBO. To be fair in our comparison, we show the results of the MMVAE using the IWAE bound $(k > 1)$ and the ELBO bound $(k=1)$. 

\subsection{Evaluation Metrics}
We use several different metrics : first the mean joint log-likelihoods $\ln p_\theta(X)$ of each model are computed using 1000 importance samples from the approximate joint posterior.
We also want to evaluate the conditional likelihoods. We generate from modality $i$ to $j$ by sampling $z\sim q_{\phi_i}(z|x_i)$ (or $q_{\phi_i}(z|g_i(x_i))$) and decoding this $z$ with $p_\theta(x_j|z)$. This defines a conditional generative model with resulting likelihood $p_{\theta, \phi_j}(x_i|x_j) \vcentcolon= \int_z p_\theta(x_i|z)q_{\phi_j}(z|x_j)dz$. We compute Monte-Carlo estimates using 1000 samples. 
This expression of the conditional likelihood is different than the one that was used in previous articles \cite{suzuki2016joint, wu2018multimodal,shi2019variational} that approximated $p_\theta(x_i|x_j)$. The latter seems less relevant as it does not reflect the quality of the cross-modal generation that is performed by sampling from $p_{\theta, \phi_j}(x_i|x_j)$.

We also evaluate the \textit{coherence} of cross-modal generations using pre-trained classifiers. For each image of each modality in the test dataset, we sample from the conditional distributions in the other modalities and check that the predicted label of the generation matches the original label of the image. 
Finally, we evaluate the diversity by computing the Frechet Inception Distance ($\mathrm{FID}$) \cite{heusel_gans_2017} on the conditional generations. 

\subsection{Results on MNIST-SVHN}
\begin{figure}[h!]
    \centering
    \includegraphics[width = 0.5\linewidth]{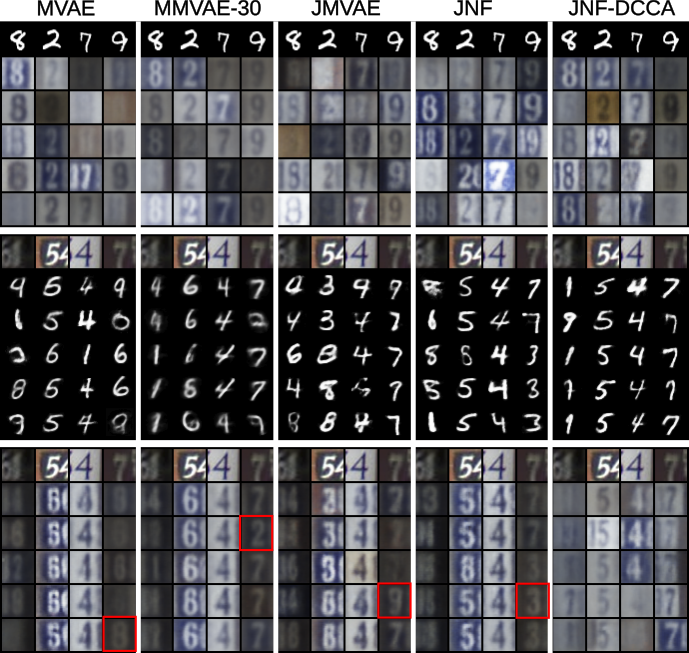}
    \caption{Samples from the conditional distributions. The first row in each image gives the samples we condition on, and the following rows are generated samples. We select in red some examples where the background is well reconstructed but not the digit. That is avoided with the DCCA.}
    \label{samples ms}
\end{figure}

The coherence and diversity results are presented in Table~\ref{validate_ms}. Our models have excellent coherence values while having the lowest $\mathrm{FID}$. The MMVAE-(k=30) has a better precision when sampling SVHN images from MNIST images but this comes at the cost of an important loss in the diversity of the generation that is reflected in the $\mathrm{FID}$s value.
Generating MNIST from SVHN is a harder task than the opposite since SVHN images are noisier and have a wider diversity (colors and backgrounds). For most methods but the JNF-DCCA and MMVAE, we observe that the latent code inferred by $q_{\phi_2}(z|x_2)$  is more influenced by the background than the digit information (see Figure.~\ref{samples ms}). 
With our JNF-DCCA method, the background information is tuned out by the DCCA, therefore the latent code inferred by $q_{\phi_2}(z|g_2(x_2))$ is based on the digit information only. As such, the conditional generation from SVHN to MNIST is much more coherent than all previous methods as reflected in Table~\ref{validate_ms}.
\begin{table}[t]
\caption{Coherence and $\mathrm{FID}$ values on the test set averaged on 5 independent runs. For $i=1,2$, $P_i$ (resp $F_i$) is the coherence (resp the $\mathrm{FID}$) when generating modality $i$ from the other. The standard deviations are all $ \leq 0.003$ for the precision values and $\leq 0.5$ for the $\mathrm{FID}$ values. The $\mathrm{FID}$ are computed on the test dataset : $\approx 50 000$ in MNIST-SVHN and $\approx 20 000$ for CelebA.}
\label{validate_ms}
\vskip 0.15in
\begin{center}
\begin{scriptsize}
\begin{sc}
\begin{tabular}{l|ccccc}
\toprule
&Model & $P_1$ & $P_2$ & $F_1$ & $F_2$ \\
\midrule
\multirow{6}{*}{\rotatebox[origin=c]{90}{MNIST-SVHN}}&MMVAE - $k=30$   & $0.606$ &$\textbf{ 0.871}$ &  $42.7$ & $104.9$  \\
&MMVAE - $k=1$ & $0.398$ & $0.140$ & $399.6$ & $106.1$ \\
&MVAE & $0.158$& $0.692$& $28.3$ & $115.6$\\
&JMVAE    & $0.468$ & $0.795$ & $13.0$ & $72.0$ \\
&JNF (Ours)    & $0.579$ & $0.834$ &    $10.6$ & $\textbf{65.5 }$    \\
&JNF-DCCA (Ours)     &$ \textbf{0.792}$& $0.811$ & $\textbf{10.3}$ & $67.4$\\
\midrule
\parbox[t]{2mm}{\multirow{6}{*}{\rotatebox[origin=c]{90}{CELEBA}}}& MMVAE - $k=15$  & $0.845$ &$\textbf{ 0.874}$ & $121.5$& /\\
&MMVAE - $k=1$  & $\textbf{0.851}$ & $\textbf{0.893}$& $153.3$&/\\
&MVAE & $0.823$ & $0.799$& $78.9$&/ \\
&JMVAE    & $0.825$ & $0.867$ & $64.6$&/ \\
&JNF (Ours)    & $0.841$ & $0.864$ &    $\textbf{62.7}$    &/ \\
&JNF-DCCA  (Ours)   & $0.844$& $0.857$ & \textbf{62.5}&/ \\
\bottomrule
\end{tabular}
\end{sc}
\end{scriptsize}
\end{center}
\vskip -0.1in
\end{table}
Hence our models reach the lowest $\mathrm{FID}$ and the highest $P_1$ value while having a good $P_2$ coherence. Table \ref{lik_ms} shows that our models also reach \textit{state-of-the-art} likelihoods for the conditional distributions and second best joint likelihood.   
\begin{table}[t]
\caption{Conditional and joint log-likelihoods averaged over the test dataset containing 50000 samples for MNIST-SVHN and 20000 for CelebA. For fair comparison, we do not compare our models to the MMVAE using IWAE, known to give higher likelihoods than the ELBO, and only provide the results as an indication.}
\label{lik_ms}
\vskip 0.15in
\begin{center}
\begin{scriptsize}
\begin{sc}
\begin{tabular}{l|cccc}
\toprule
&Model & $\ln p(x_2|x_1) $ & $\ln p(x_1|x_2) $ & $\ln p(x_1,x_2)$\\
\midrule
\parbox[t]{2mm}{\multirow{6}{*}{\rotatebox[origin=c]{90}{MNIST-SVHN}}}&\emph{MMVAE - $k=30$  }    & $-\emph{2848}$  & $-\emph{738}$ & $-\emph{3594}$\\
&MMVAE - $k=1$ & $-2898$ & $-742$ & $-3599$\\
&MVAE & $-2847$ & $-741$ & $\mathbf{-3586}$\\
&JMVAE    & $-2847$ & $-741$ & $-3590$\\
&JNF (Ours)   & $\mathbf{-2846}$ & $-741$ & $-3590$\\
&JNF-DCCA (Ours)     & $\mathbf{-2846}$ &$\textbf{-739}$ & $-3590$\\
\midrule
\parbox[t]{2mm}{\multirow{6}{*}{\rotatebox[origin=c]{90}{CELEBA}}}&\emph{MMVAE - $k=15$  } & $-\textit{11477}$ & $-\textit{9.0}$ & $-\emph{11476}$\\
&MMVAE - $k=1$   & $-11607$ & $\mathbf{-9.4}$ & $-11489$\\
&MVAE & $-11503$ & $-21.8$ & $-11418$ \\
&JMVAE& $-11490$ & $-10.1$ & $\mathbf{-11414}$\\
&JNF (Ours)& $-11482$ & $-10.1$ & $\mathbf{-11414}$\\
&JNF-DCCA (Ours)& $\mathbf{-11481}$& $-10.1$& $\mathbf{-11414}$\\
\bottomrule
\end{tabular}
\end{sc}
\end{scriptsize}
\end{center}
\vskip -0.1in
\end{table}



\subsection{Results on CelebA}



\begin{figure}[h!]
    \centering
    \includegraphics[width=0.7\linewidth]{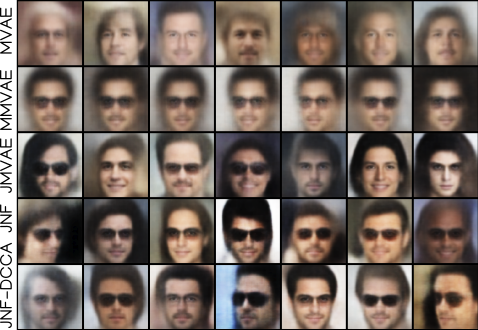}
    \caption{Some generated samples conditioned on a vector of attributes containing "Black Hair, Eyeglasses,Goatee, Male, Mouth-Slightly-Open, Mustache, Young". The complete set of attributes for this generated samples are in Appendix.\ref{app:more celeba samples}.}
    \label{samples celeba}
\end{figure}

Table~\ref{validate_ms} shows that our models (JNF, JNF-DCCA) have good coherence values while having the lowest $\mathrm{FID}$ values. The MMVAE obtain slightly better coherences but at the cost of an important lack of diversity in the generated images (see  Figure.\ref{samples celeba} and $\mathrm{FID}$s values).
On this dataset, we do not observe a difference when using the DCCA since the shared information between the modalities is exactly the second modality (the attributes). 
Nevertheless, we see that using it does not diminish the performance of the method: the accuracy is slightly reduced when predicting the attributes from the images but slightly heightened the other way around.
Table~\ref{lik_ms} shows that our methods outperform other models except the MMVAE on one conditional distribution.
Qualitative samples are also presented in Figure.\ref{samples celeba}.

\subsection{Results on a Trimodal Dataset}
\begin{table*}[ht]
\caption{Coherences and likelihoods for the MNIST-SVHN-FashionMNIST dataset. The coherences are averaged on 5 independent runs with a standard deviation $\leq 0.003$ for all results. The likelihoods are averaged  on the test set containing $\approx 50000$ samples.}

\label{msf validate}
\vskip 0.15in
\begin{center}
\begin{scriptsize}
\begin{sc}
\begin{tabular}{l|cccccccccc}
\toprule
&\multirow{2}{*}{Model}&S&F&S,F&M&F&M,F&M&S&M,S\\
 &&M&M&M&S&S&S&F&F&F\\
\midrule
\multirow{5}{*}{\rotatebox[origin=c]{90}{ Coherences}}&MMVAE - $k=30$& $0.722$ & $\mathbf{0.832}$ & $0.777$ &  $0.754$ & $0.677$ & $0.715$ & $0.810$ & $0.635$ & $0.723$\\
&MMMVAE - $k=1$& $0.563$& $0.718$ & $0.640$ & $0.793$ & $0.705$ & $0.749$ & $\mathbf{0.814}$ & $0.618$ & $0.716$\\
&MVAE & $0.170$& $0.170$ & $0.168$ & $0.532$ & $0.349$ & $0.427$ &$0.171$& $0.148$ & $0.157$ \\
&JMVAE & $0.530$ & $0.703$   & $0.804$ & $0.757$ & 0$.654$ & $0.845$ & $0.724$& $0.492$ & $0.785$ \\
&JNF    &$0.560$ & $0.732$ & $0.840$& $0.797$ & $0.688$ & $0.879$ & $0.758$& $0.494$ & $0.819$\\
&JNF-DCCA     &$\mathbf{0.765}$ &$0.801$& $\mathbf{0.876}$
 & $\mathbf{0.820}$& $\mathbf{0.749}$ & $\mathbf{0.880}$ & $0.811$ & $\mathbf{0.698}$ &$\mathbf{0.842 }$\\

\end{tabular}

\begin{tabular}{l|cccccccr}
\toprule
&Model &$P(M|S)$&$P(M,F)$&$P(S|M)$&$P(S|F)$&$P(F|M)$&$P(F|S)$&$P(M,S,F)$\\
\midrule
\multirow{5}{*}{\rotatebox[origin=c]{90}{Likelihoods}}&\textit{MMVAE - $k=30$}& $-\textit{738.7}$ & $-\textit{738.0}$ &  $-\textit{2853.7}$ &$-\textit{2852.3}$ &$-\textit{733.8}$ & $-\textit{734.7}$  & $-\textit{4338}$\\
&MMVAE - $k=1$& $-742.5$& $-742.1$ & $-2898.3$ & $-2897.0$ & $-740.7$ & $-742.4$ & $-4349$\\

&MVAE & $-742.5$ & $-741.6$ & $-2847.6$ & $-2847.7$  & $-736.5$ & $-737.0$ & $-4325$\\
&JMVAE & $-738.3$ & $-738.0$   &$-2847.1$ & $-2847.4$  & $-734.2$ & $-734.7$ &$\mathbf{-4318}$\\
&JNF    & $\mathbf{- 737.8}$& $-737.2$ & $\mathbf{-2846.9 }$& $-2847.1$ & $\mathbf{-733.5}$ & $\mathbf{-734.2}$ &$\mathbf{ -4318}$ \\
&JNF-DCCA     & $\mathbf{-737.8 }$&$\textbf{-737.1}$& $\mathbf{-2846.9 }$&$\mathbf{ -2847.0}$ &$\mathbf{-733.5}$ & $-735.0$ & $\mathbf{-4318}$ \\

\bottomrule
\end{tabular}

\end{sc}
\end{scriptsize}
\end{center}
\vskip -0.1in
\end{table*}

Finally, we demonstrate the scalability of our method on a trimodal dataset. 
With this trimodal dataset, we want to evaluate the joint distribution and the generations of modality $i$ conditioned on a subset $S$ of $[|1,3|]$, $p_{\theta}(x_i|x_{j \in S})$.
For $|S| >1$, the MVAE and MMVAE models include ways to sample from those by modelling them as the aggregation of the unimodal inference distributions (with either a Product of Experts or a Mixture of Experts). The JMVAE does not include a way to handle that case but we propose to use a PoE of the unimodal encoders as in Eq.~\eqref{why poe}. For the JMVAE, JNF and JNF-DCCA models we propose sampling from the PoE using Hamiltonian Monte Carlo Sampling.  
Table.\ref{msf validate} shows that our methods reach the best coherences, and likelihoods for almost all the distributions. They obtain especially good coherences when conditioning on a subset even though we do not specifically train for this scenario. Using the DCCA results in a significant gain in accuracy. The MMVAE model also reach good coherences but Appendix.~\ref{add results on msf} shows that the images generated by our model are much more diverse.

\section{Conclusion and Perspectives}

In this paper, we have introduced two new multimodal variational autoencoders that integrate Normalizing Flows and conditioning on DCCA embeddings. We demonstrate on a toy dataset how the Normalizing Flows allow to better fit the unimodal posteriors and therefore improve the coherence of the conditional generations. The relevance of using flows and DCCA embeddings is demonstrated on three benchmark datasets. In particular, we observe a significant gain in coherence in the conditional distributions for the MNIST-SVHN and the MNIST-SVHN-FashionMNIST datasets. The latter shows that using a Product of Experts of the unimodal posteriors at inference time is extremely relevant for sampling from the subset posteriors. 
The general DCCA embedding used in this paper might be replaced by another method more specific to the type of data that is used to improve the results. This could be investigated in future work.



\clearpage
\clearpage
\bibliography{MyLibrary}

\begin{thebibliography}{50}
\providecommand{\natexlab}[1]{#1}
\providecommand{\url}[1]{\texttt{#1}}
\expandafter\ifx\csname urlstyle\endcsname\relax
  \providecommand{\doi}[1]{doi: #1}\else
  \providecommand{\doi}{doi: \begingroup \urlstyle{rm}\Url}\fi

\bibitem[Andrew et~al.(2013)Andrew, Arora, Bilmes, and Livescu]{andrew2013deep}
Andrew, G., Arora, R., Bilmes, J., and Livescu, K.
\newblock Deep canonical correlation analysis.
\newblock In \emph{International conference on machine learning}, pp.\
  1247--1255. PMLR, 2013.

\bibitem[Biewald(2020)]{wandb}
Biewald, L.
\newblock Experiment tracking with weights and biases, 2020.
\newblock URL \url{https://www.wandb.com/}.
\newblock Software available from wandb.com.

\bibitem[Burda et~al.(2015)Burda, Grosse, and
  Salakhutdinov]{burda2015importance}
Burda, Y., Grosse, R., and Salakhutdinov, R.
\newblock Importance weighted autoencoders.
\newblock \emph{arXiv preprint arXiv:1509.00519}, 2015.

\bibitem[Chadebec et~al.(2022{\natexlab{a}})Chadebec, Thibeau-Sutre, Burgos,
  and Allassonnière]{chadebec_data_2022}
Chadebec, C., Thibeau-Sutre, E., Burgos, N., and Allassonnière, S.
\newblock Data {Augmentation} in {High} {Dimensional} {Low} {Sample} {Size}
  {Setting} {Using} a {Geometry}-{Based} {Variational} {Autoencoder}.
\newblock \emph{IEEE Transactions on Pattern Analysis and Machine
  Intelligence}, pp.\  1--18, 2022{\natexlab{a}}.
\newblock ISSN 1939-3539.
\newblock \doi{10.1109/TPAMI.2022.3185773}.
\newblock Conference Name: IEEE Transactions on Pattern Analysis and Machine
  Intelligence.

\bibitem[Chadebec et~al.(2022{\natexlab{b}})Chadebec, Vincent, and
  Allassonniere]{chadebecpythae}
Chadebec, C., Vincent, L.~J., and Allassonniere, S.
\newblock Pythae: Unifying generative autoencoders in python-a benchmarking use
  case.
\newblock In \emph{Thirty-sixth Conference on Neural Information Processing
  Systems Datasets and Benchmarks Track}, 2022{\natexlab{b}}.

\bibitem[Daunhawer et~al.(2021)Daunhawer, Sutter, Marcinkevi{\v{c}}s, and
  Vogt]{daunhawer2021self}
Daunhawer, I., Sutter, T.~M., Marcinkevi{\v{c}}s, R., and Vogt, J.~E.
\newblock Self-supervised disentanglement of modality-specific and shared
  factors improves multimodal generative models.
\newblock In \emph{Pattern Recognition: 42nd DAGM German Conference, DAGM GCPR
  2020, T{\"u}bingen, Germany, September 28--October 1, 2020, Proceedings 42},
  pp.\  459--473. Springer, 2021.

\bibitem[Daunhawer et~al.(2022)Daunhawer, Sutter, Chin-Cheong, Palumbo, and
  Vogt]{daunhawer_limitations_2022}
Daunhawer, I., Sutter, T.~M., Chin-Cheong, K., Palumbo, E., and Vogt, J.~E.
\newblock On the limitations of multimodal vaes.
\newblock In \emph{International Conference on Learning Representations}, 2022.

\bibitem[Duane et~al.(1987)Duane, Kennedy, Pendleton, and
  Roweth]{duane_hybrid_1987}
Duane, S., Kennedy, A.~D., Pendleton, B.~J., and Roweth, D.
\newblock Hybrid monte carlo.
\newblock \emph{Physics Letters B}, 195\penalty0 (2):\penalty0 216--222, 1987.

\bibitem[Girolami \& Calderhead(2011)Girolami and
  Calderhead]{girolami_riemann_2011}
Girolami, M. and Calderhead, B.
\newblock Riemann manifold langevin and hamiltonian monte carlo methods.
\newblock \emph{Journal of the Royal Statistical Society: Series B (Statistical
  Methodology)}, 73\penalty0 (2):\penalty0 123--214, 2011.

\bibitem[Goodfellow et~al.(2014)Goodfellow, Pouget-Abadie, Mirza, Xu,
  Warde-Farley, Ozair, Courville, and Bengio]{goodfellow_generative_2014}
Goodfellow, I., Pouget-Abadie, J., Mirza, M., Xu, B., Warde-Farley, D., Ozair,
  S., Courville, A., and Bengio, Y.
\newblock Generative {Adversarial} {Nets}.
\newblock In \emph{Advances in {Neural} {Information} {Processing} {Systems}},
  volume~27. Curran Associates, Inc., 2014.

\bibitem[Hardoon et~al.(2004)Hardoon, Szedmak, and
  Shawe-Taylor]{hardoon_canonical_2004}
Hardoon, D.~R., Szedmak, S., and Shawe-Taylor, J.
\newblock Canonical correlation analysis: an overview with application to
  learning methods.
\newblock \emph{Neural Computation}, 16\penalty0 (12):\penalty0 2639--2664,
  December 2004.
\newblock ISSN 0899-7667.
\newblock \doi{10.1162/0899766042321814}.

\bibitem[He et~al.(2016)He, Zhang, Ren, and Sun]{he2016deep}
He, K., Zhang, X., Ren, S., and Sun, J.
\newblock Deep residual learning for image recognition.
\newblock In \emph{Proceedings of the IEEE conference on computer vision and
  pattern recognition}, pp.\  770--778, 2016.

\bibitem[Heusel et~al.(2017)Heusel, Ramsauer, Unterthiner, Nessler, and
  Hochreiter]{heusel_gans_2017}
Heusel, M., Ramsauer, H., Unterthiner, T., Nessler, B., and Hochreiter, S.
\newblock Gans trained by a two time-scale update rule converge to a local nash
  equilibrium.
\newblock In \emph{Advances in {Neural} {Information} {Processing} {Systems}},
  2017.

\bibitem[Hsu \& Glass(2018)Hsu and Glass]{hsu2018disentangling}
Hsu, W.-N. and Glass, J.
\newblock Disentangling by partitioning: A representation learning framework
  for multimodal sensory data.
\newblock \emph{arXiv preprint arXiv:1805.11264}, 2018.

\bibitem[Isola et~al.(2017)Isola, Zhu, Zhou, and Efros]{isola2017image}
Isola, P., Zhu, J.-Y., Zhou, T., and Efros, A.~A.
\newblock Image-to-image translation with conditional adversarial networks.
\newblock In \emph{Proceedings of the IEEE conference on computer vision and
  pattern recognition}, pp.\  1125--1134, 2017.

\bibitem[Javaloy et~al.(2022)Javaloy, Meghdadi, and
  Valera]{javaloy2022mitigating}
Javaloy, A., Meghdadi, M., and Valera, I.
\newblock Mitigating modality collapse in multimodal vaes via impartial
  optimization.
\newblock In \emph{International Conference on Machine Learning}, pp.\
  9938--9964. PMLR, 2022.

\bibitem[Jordan et~al.(1999)Jordan, Ghahramani, Jaakkola, and
  Saul]{jordan1999introduction}
Jordan, M.~I., Ghahramani, Z., Jaakkola, T.~S., and Saul, L.~K.
\newblock An introduction to variational methods for graphical models.
\newblock \emph{Machine learning}, 37:\penalty0 183--233, 1999.

\bibitem[Kanatsoulis et~al.(2018)Kanatsoulis, Fu, Sidiropoulos, and
  Hong]{kanatsoulis2018structured}
Kanatsoulis, C.~I., Fu, X., Sidiropoulos, N.~D., and Hong, M.
\newblock Structured sumcor multiview canonical correlation analysis for
  large-scale data.
\newblock \emph{IEEE Transactions on Signal Processing}, 67\penalty0
  (2):\penalty0 306--319, 2018.

\bibitem[Kingma \& Welling(2013)Kingma and Welling]{kingma2013auto}
Kingma, D.~P. and Welling, M.
\newblock Auto-encoding variational bayes.
\newblock \emph{arXiv preprint arXiv:1312.6114}, 2013.

\bibitem[Korthals et~al.(2019)Korthals, Rudolph, Leitner, Hesse, and
  R{\"u}ckert]{korthals2019multi}
Korthals, T., Rudolph, D., Leitner, J., Hesse, M., and R{\"u}ckert, U.
\newblock Multi-modal generative models for learning epistemic active sensing.
\newblock In \emph{2019 International Conference on Robotics and Automation
  (ICRA)}, pp.\  3319--3325. IEEE, 2019.

\bibitem[Lecun et~al.(1998)Lecun, Bottou, Bengio, and
  Haffner]{lecun_gradient-based_1998}
Lecun, Y., Bottou, L., Bengio, Y., and Haffner, P.
\newblock Gradient-based learning applied to document recognition.
\newblock \emph{Proceedings of the IEEE}, 86\penalty0 (11):\penalty0
  2278--2324, November 1998.
\newblock ISSN 1558-2256.
\newblock \doi{10.1109/5.726791}.
\newblock Conference Name: Proceedings of the IEEE.

\bibitem[Li et~al.(2020)Li, Yu, Wang, and Zheng]{li2020tumorgan}
Li, Q., Yu, Z., Wang, Y., and Zheng, H.
\newblock Tumorgan: A multi-modal data augmentation framework for brain tumor
  segmentation.
\newblock \emph{Sensors}, 20\penalty0 (15):\penalty0 4203, 2020.

\bibitem[Liu(2008)]{liu_monte_2008}
Liu, J.~S.
\newblock \emph{Monte Carlo strategies in scientific computing}.
\newblock Springer Science \& Business Media, 2008.

\bibitem[Liu et~al.(2015)Liu, Luo, Wang, and Tang]{liu2015deep}
Liu, Z., Luo, P., Wang, X., and Tang, X.
\newblock Deep learning face attributes in the wild.
\newblock In \emph{Proceedings of the IEEE international conference on computer
  vision}, pp.\  3730--3738, 2015.

\bibitem[Neal(2005)]{neal2005hamiltonian}
Neal, R.~M.
\newblock Hamiltonian importance sampling.
\newblock In \emph{talk presented at the Banff International Research Station
  (BIRS) workshop on Mathematical Issues in Molecular Dynamics}, 2005.

\bibitem[Neal \& {others}(2011)Neal and {others}]{neal_mcmc_2011}
Neal, R.~M. and {others}.
\newblock {MCMC} using hamiltonian dynamics.
\newblock \emph{Handbook of Markov Chain Monte Carlo}, 2\penalty0
  (11):\penalty0 2, 2011.

\bibitem[Netzer et~al.(2011)Netzer, Wang, Coates, Bissacco, Wu, and
  Ng]{netzer_reading_2011}
Netzer, Y., Wang, T., Coates, A., Bissacco, A., Wu, B., and Ng, A.
\newblock Reading {Digits} in {Natural} {Images} with {Unsupervised} {Feature}
  {Learning}.
\newblock \emph{NIPS}, January 2011.

\bibitem[Palumbo et~al.(2022)Palumbo, Daunhawer, and Vogt]{palumbo2022mmvae+}
Palumbo, E., Daunhawer, I., and Vogt, J.~E.
\newblock Mmvae+: Enhancing the generative quality of multimodal vaes without
  compromises.
\newblock In \emph{ICLR Workshop on Deep Generative Models for Highly
  Structured Data}, 2022.

\bibitem[Papamakarios et~al.(2017)Papamakarios, Pavlakou, and
  Murray]{papamakarios2017masked}
Papamakarios, G., Pavlakou, T., and Murray, I.
\newblock Masked autoregressive flow for density estimation.
\newblock \emph{Advances in neural information processing systems}, 30, 2017.

\bibitem[Paszke et~al.(2017)Paszke, Gross, Chintala, Chanan, Yang, DeVito, Lin,
  Desmaison, Antiga, and Lerer]{paszke2017automatic}
Paszke, A., Gross, S., Chintala, S., Chanan, G., Yang, E., DeVito, Z., Lin, Z.,
  Desmaison, A., Antiga, L., and Lerer, A.
\newblock Automatic differentiation in pytorch.
\newblock In \emph{NIPS-W}, 2017.

\bibitem[Rezende \& Mohamed(2015)Rezende and Mohamed]{rezende2015variational}
Rezende, D. and Mohamed, S.
\newblock Variational inference with normalizing flows.
\newblock In \emph{International conference on machine learning}, pp.\
  1530--1538. PMLR, 2015.

\bibitem[Saxena \& Cao(2021)Saxena and Cao]{saxena2021generative}
Saxena, D. and Cao, J.
\newblock Generative adversarial networks (gans) challenges, solutions, and
  future directions.
\newblock \emph{ACM Computing Surveys (CSUR)}, 54\penalty0 (3):\penalty0 1--42,
  2021.

\bibitem[Shi et~al.(2019)Shi, Paige, Torr, et~al.]{shi2019variational}
Shi, Y., Paige, B., Torr, P., et~al.
\newblock Variational mixture-of-experts autoencoders for multi-modal deep
  generative models.
\newblock \emph{Advances in Neural Information Processing Systems}, 32, 2019.

\bibitem[Shin et~al.(2018)Shin, Tenenholtz, Rogers, Schwarz, Senjem, Gunter,
  Andriole, and Michalski]{shin2018medical}
Shin, H.-C., Tenenholtz, N.~A., Rogers, J.~K., Schwarz, C.~G., Senjem, M.~L.,
  Gunter, J.~L., Andriole, K.~P., and Michalski, M.
\newblock Medical image synthesis for data augmentation and anonymization using
  generative adversarial networks.
\newblock In \emph{Simulation and Synthesis in Medical Imaging: Third
  International Workshop, SASHIMI 2018, Held in Conjunction with MICCAI 2018,
  Granada, Spain, September 16, 2018, Proceedings 3}, pp.\  1--11. Springer,
  2018.

\bibitem[Shorten \& Khoshgoftaar(2019)Shorten and
  Khoshgoftaar]{shorten2019survey}
Shorten, C. and Khoshgoftaar, T.~M.
\newblock A survey on image data augmentation for deep learning.
\newblock \emph{Journal of big data}, 6\penalty0 (1):\penalty0 1--48, 2019.

\bibitem[Sutter et~al.(2020)Sutter, Daunhawer, and Vogt]{sutter2020multimodal}
Sutter, T., Daunhawer, I., and Vogt, J.
\newblock Multimodal generative learning utilizing jensen-shannon-divergence.
\newblock \emph{Advances in neural information processing systems},
  33:\penalty0 6100--6110, 2020.

\bibitem[Sutter et~al.(2021)Sutter, Daunhawer, and
  Vogt]{sutter_generalized_2021}
Sutter, T.~M., Daunhawer, I., and Vogt, J.~E.
\newblock Generalized {Multimodal} {ELBO}.
\newblock \emph{ICLR}, 2021.

\bibitem[Suzuki \& Matsuo(2022)Suzuki and Matsuo]{suzuki2022survey}
Suzuki, M. and Matsuo, Y.
\newblock A survey of multimodal deep generative models.
\newblock \emph{Advanced Robotics}, 36\penalty0 (5-6):\penalty0 261--278, 2022.

\bibitem[Suzuki et~al.(2016)Suzuki, Nakayama, and Matsuo]{suzuki2016joint}
Suzuki, M., Nakayama, K., and Matsuo, Y.
\newblock Joint multimodal learning with deep generative models.
\newblock \emph{arXiv preprint arXiv:1611.01891}, 2016.

\bibitem[Tanner \& Wong(1987)Tanner and Wong]{tanner1987calculation}
Tanner, M.~A. and Wong, W.~H.
\newblock The calculation of posterior distributions by data augmentation.
\newblock \emph{Journal of the American statistical Association}, 82\penalty0
  (398):\penalty0 528--540, 1987.

\bibitem[Tian et~al.(2020)Tian, Krishnan, and Isola]{tian2020contrastive}
Tian, Y., Krishnan, D., and Isola, P.
\newblock Contrastive multiview coding.
\newblock In \emph{Computer Vision--ECCV 2020: 16th European Conference,
  Glasgow, UK, August 23--28, 2020, Proceedings, Part XI 16}, pp.\  776--794.
  Springer, 2020.

\bibitem[Van~der Maaten \& Hinton(2008)Van~der Maaten and
  Hinton]{van2008visualizing}
Van~der Maaten, L. and Hinton, G.
\newblock Visualizing data using t-sne.
\newblock \emph{Journal of machine learning research}, 9\penalty0 (11), 2008.

\bibitem[Vedantam et~al.(2018)Vedantam, Fischer, Huang, and
  Murphy]{vedantamgenerative}
Vedantam, R., Fischer, I., Huang, J., and Murphy, K.
\newblock Generative models of visually grounded imagination.
\newblock In \emph{International Conference on Learning Representations}, 2018.

\bibitem[Wang et~al.(2016)Wang, Yan, Lee, and Livescu]{wang2016deep}
Wang, W., Yan, X., Lee, H., and Livescu, K.
\newblock Deep variational canonical correlation analysis.
\newblock \emph{arXiv preprint arXiv:1610.03454}, 2016.

\bibitem[Wei et~al.(2019)Wei, Suriawinata, Vaickus, Ren, Liu, Wei, and
  Hassanpour]{wei2019generative}
Wei, J., Suriawinata, A., Vaickus, L., Ren, B., Liu, X., Wei, J., and
  Hassanpour, S.
\newblock Generative image translation for data augmentation in colorectal
  histopathology images.
\newblock \emph{Proceedings of machine learning research}, 116:\penalty0 10,
  2019.

\bibitem[Wu \& Goodman(2018{\natexlab{a}})Wu and Goodman]{wu2018multimodal}
Wu, M. and Goodman, N.
\newblock Multimodal generative models for scalable weakly-supervised learning.
\newblock \emph{Advances in neural information processing systems}, 31,
  2018{\natexlab{a}}.

\bibitem[Wu \& Goodman(2018{\natexlab{b}})Wu and Goodman]{wu_multimodal_2018}
Wu, M. and Goodman, N.
\newblock Multimodal {Generative} {Models} for {Scalable} {Weakly}-{Supervised}
  {Learning}.
\newblock In \emph{Advances in {Neural} {Information} {Processing} {Systems}},
  volume~31. Curran Associates, Inc., 2018{\natexlab{b}}.
\newblock URL
  \url{https://proceedings.neurips.cc/paper/2018/hash/1102a326d5f7c9e04fc3c89d0ede88c9-Abstract.html}.

\bibitem[Xiao et~al.(2017)Xiao, Rasul, and Vollgraf]{xiao2017fashion}
Xiao, H., Rasul, K., and Vollgraf, R.
\newblock Fashion-{MNIST}: a novel image dataset for benchmarking machine
  learning algorithms.
\newblock \emph{arXiv preprint arXiv:1708.07747}, 2017.

\bibitem[Yin et~al.(2017)Yin, Melo, Billard, and Paiva]{yin2017associate}
Yin, H., Melo, F., Billard, A., and Paiva, A.
\newblock Associate latent encodings in learning from demonstrations.
\newblock In \emph{Proceedings of the AAAI Conference on Artificial
  Intelligence}, volume~31, 2017.

\bibitem[Zhu et~al.(2017)Zhu, Zhang, Pathak, Darrell, Efros, Wang, and
  Shechtman]{zhu2017toward}
Zhu, J.-Y., Zhang, R., Pathak, D., Darrell, T., Efros, A.~A., Wang, O., and
  Shechtman, E.
\newblock Toward multimodal image-to-image translation.
\newblock \emph{Advances in neural information processing systems}, 30, 2017.

\end{thebibliography}
\bibliographystyle{icml2023}

\clearpage
\appendix
\onecolumn

\section{Interpretations of the $\mathcal{L}_{\mathrm{JM}}$ Objective}
\label{interpretations}
In this appendix, we provide several interpretations of the $\mathcal{L}_{\mathrm{JM}}$ (Eq.~\eqref{ljm} and Eq.~\eqref{rewrite_ljm}) that explains why minimizing it is a sensible objective to fit the unimodal posteriors. Firstly, we recall an analysis from \cite{suzuki2016joint} that links $\mathcal{L}_{\mathrm{JM}}$ to the notion of Variation of Information. Secondly, we reinterpret $\mathcal{L}_{\mathrm{JM}}$ to show that it brings the unimodal encoder $q_{\phi_i}(z|x_i)$ (for $i\in[1,m]$) close to an average distribution $q_{\mathrm{avg}}(z|x_i) =  \mathbb{E}_{\hat{p}((x_j)_{j\neq i}|x_i)}(q_\phi(z|X))$ that is close to $ p_\theta(z|x_i)$ provided that the joint encoder is well fit. 

\subsection{Interpretation in Relation to the Variation of Information}
First, in the bimodal case, we recall an interpretation provided by \cite{suzuki2016joint} that links $\mathcal{L}_{JM}$ to the Variation of Information (VI) of $x_1$ and $x_2$ where $x_1$ (resp. $x_2$) represent the variable of the first modality (resp second).
Recall the definition of the VI : 
\begin{equation}
\label{vi def}
VI(x_1,x_2) = - \mathbb{E}_{\mathbb{P}(x_1,x_2)} \big(\ln \mathbb{P}(x_1|x_2) + \ln \mathbb{P}(x_2|x_1) \big)\,.
\end{equation}
If we analyse Eq.~\eqref{vi def}, we see that the more the modalities are predictive of one another, the smaller is the Variation of Information. 
However, we do not know the true joint and conditional distributions but we can use the following approximation summing on $N$ training samples:
\begin{equation*}
    \widetilde{VI} = - \sum_{n=1}^N \ln p_{\theta,\phi_1}(x_1^{(n)}|x_2^{(n)}) + \ln p_{\theta, \phi_2}(x_2^{(n)}|x_1^{(n)})\,,
\end{equation*}
where for $i,j \in \{1,2\}$ with $i\neq j$, $p_{\theta,\phi_i}(x_j|x_i) \vcentcolon= \int p_\theta(x_j|z) q_{\phi_i}(z|x_j)dz$ is our conditional generative models to sample $x_j$ from $x_i$. 
We can show that with $\mathcal{L}$ being the ELBO defined in Eq.~\eqref{elbo def} and $\mathcal{L}_{JM}$ defined in Eq.~\eqref{ljm}:
\begin{equation}
\label{bound on vi}
    - \mathcal{L}(x_1,x_2) + \mathcal{L}_{\mathrm{JM}}(x_1,x_2) \geq \widetilde{VI}\,.
\end{equation}
We recall that in our method, we minimize $\mathcal{L}_{\mathrm{JM}}(x_1,x_2)$ during the second step of our training with $\mathcal{L}(x_1,x_2)$ fixed, therefore we minimize an upper bound on $\Tilde{VI}$ that is the empirical Variation of Information between modality 1 and 2. Minimizing $\Tilde{VI}$ is a sensible goal as it encapsulates the predictive power of a modality given the other.

Let us now prove Eq.~\eqref{bound on vi} :
\begin{equation*}
    \begin{split}
        \ln p_{\theta,\phi_1}(x_2|x_1) + \ln p_{\theta,\phi_2}(x_1|x_2) &\geq \mathbb{E}_{q_\phi(z|x_1,x_2)} \left ( \ln \frac{p_\theta(x_1|z)q_{\phi_2}(z|x_2)}{q_\phi(z|x_1,x_2)} \right ) + \mathbb{E}_{q_\phi(z|x_1,x_2)}\big( \ln \frac{p_\theta(x_2|z)q_{\phi_1}(z|x_1)}{q_\phi(z|x_1,x_2)} \big)\\
        &= \mathbb{E}_{q_\phi(z|x_1,x_2)}\big(\ln p_\theta(x_1|z)) + \mathbb{E}_{q_\phi(z|x_1,x_2)} \big(\ln p_\theta(x_2|z) \big )\\ 
        &- KL(q_\phi(z|x_1,x_2) || q_{\phi_2}(z|x_2)) - KL(q_\phi(|x_1,x_2) || q_{\phi_1}(z|x_1))\\
        &= \mathcal{L}(x_1,x_2) + KL(q_\phi(z|x_1,x_2) ||p(z)) - \mathcal{L}_{\mathrm{JM}}(x_1,x_2)\,.
    \end{split}
\end{equation*}

\subsection{Interpretation in Relation to an Average Distribution}
In a second time, we provide an interpretation inspired by \cite{vedantamgenerative} but extended to our case with continuous variables. We consider the second step of our training process with the joint encoder $q_{\phi}(z|X)$ fixed. Then, we integrate the expression of $\mathcal{L}_{\mathrm{JM}}(X)$ given by Eq.~\eqref{rewrite_ljm} over the empirical data distribution $\hat{p}(X)$:
\begin{equation}
\begin{split}
    \mathbb{E}_{\hat{p}(X)}(\mathcal{L}_{\mathrm{JM}}(X)) &= \sum_{i=1}^{m} \mathbb{E}_{\hat{p}(X)}\left(\mathbb{E}_{q_\phi(z|X)}\left(-\ln q_{\phi_i}(z|x_i) \right)\right)\\
     &= \sum_{i=1}^{m} \mathbb{E}_{\hat{p}(x_i)}\left(\mathbb{E}_{\hat{p}((x_j)_{j\neq i}|x_i)}(\mathbb{E}_{q_\phi(z|X)}\left(-\ln q_{\phi_i}(z|x_i) \right)\right)\,.
\end{split}
\end{equation}
If we suppose that for all $i \in [|1,m|]$, $q_{\phi_i}(z|x_i)$ is bounded by $C$, then we can continue with:
\begin{equation}
\begin{split}
    \mathbb{E}_{\hat{p}(x)}(\mathcal{L}_{\mathrm{JM}}(X)) &= \sum_{i=1}^{m} \mathbb{E}_{\hat{p}(x_i)}\left(\mathbb{E}_{\hat{p}((x_j)_{j\neq i}|x_i)}\left(\mathbb{E}_{q_\phi(z|X)}\left(-\ln \frac{q_{\phi_i}(z|x_i)}{C} \right)\right)\right) - \ln(C)\,.
\end{split}
\end{equation}
Since $-\ln \frac{q_{\phi_i}(z|x_i)}{C}$ is always positive we use Fubini's Theorem to exchange the expectations on $z$ and the $(x_j)_{j\neq i}$:

\begin{equation}
\begin{split}
    \mathbb{E}_{\hat{p}(x)}(\mathcal{L}_{\mathrm{JM}}(X)) &= \sum_{i=1}^{m} \mathbb{E}_{\hat{p}(x_i)} \int_z \int_{(x_j)_{j\neq i}} -\ln \frac{q_{\phi_i}(z|x_i)}{C} q_\phi(z|X) \hat{p}((x_j)_{j\neq i}|x_i)dz(dx_i)_{i\neq j} + cte\\
     &= \sum_{i=1}^{m} \mathbb{E}_{\hat{p}(x_i)} \int_z  -\ln \frac{q_{\phi_i}(z|x_i)}{C} \int_{(x_j)_{j\neq i}|x_i)}q_\phi(z|X) \hat{p}((x_j)_{j\neq i}|x_i)dz(dx_i)_{i\neq j} + cte\\
    &= \sum_{i=1}^{m} \mathbb{E}_{\hat{p}(x_i)}\left( \mathbb{E}_{q_{\mathrm{avg}}(z|x_i)}( -\frac{q_{\phi_i}(z|x_i)}{C}\right) +cte\\
    &= \sum_{i=1}^{m} \mathbb{E}_{\hat{p}(x_i)}\left(KL(q_{\mathrm{avg}}(z|x_i)||q_{\phi_i}(z|x_i)) + H(q_{\mathrm{avg}}(z|x_i))\right) + cte\,.
\end{split}
\end{equation}

where $q_{\mathrm{avg}}(z|x_i) =  \mathbb{E}_{\hat{p}((x_j)_{j\neq i}|x_i)}(q_\phi(z|X))$ and $H$ is the Shannon entropy. Since the entropy term does not depend on the unimodal encoders $q_{\phi_i}$, this term does not impact the training. Therefore we see that the incentive for $q_{\phi_i}(z|x_i)$ is to minimize the Kullback-Leibler divergence with $q_{\mathrm{avg}}(z|x_i)$. Getting closer to $q_{\mathrm{avg}}(z|x_i)$ is a sensible objective since, if $q_\phi(z|X)$ approximates well the true posterior $p_\theta(z|X)$ then $q_{\mathrm{avg}}(z|x_i) = \mathbb{E}_{\hat{p}((x_j)_{j\neq i}|x_i)}(q_\phi(z|X)) \approx \mathbb{E}_{\hat{p}((x_j)_{j\neq i}|x_i)}(p_\theta(z|X)) \approx p_\theta(z|x_i)$.

\clearpage

\section{On the DCCA Embeddings}
\label{app:dcca_embeddings}

\begin{figure}[ht]
    \centering
    \includegraphics[width=\linewidth]{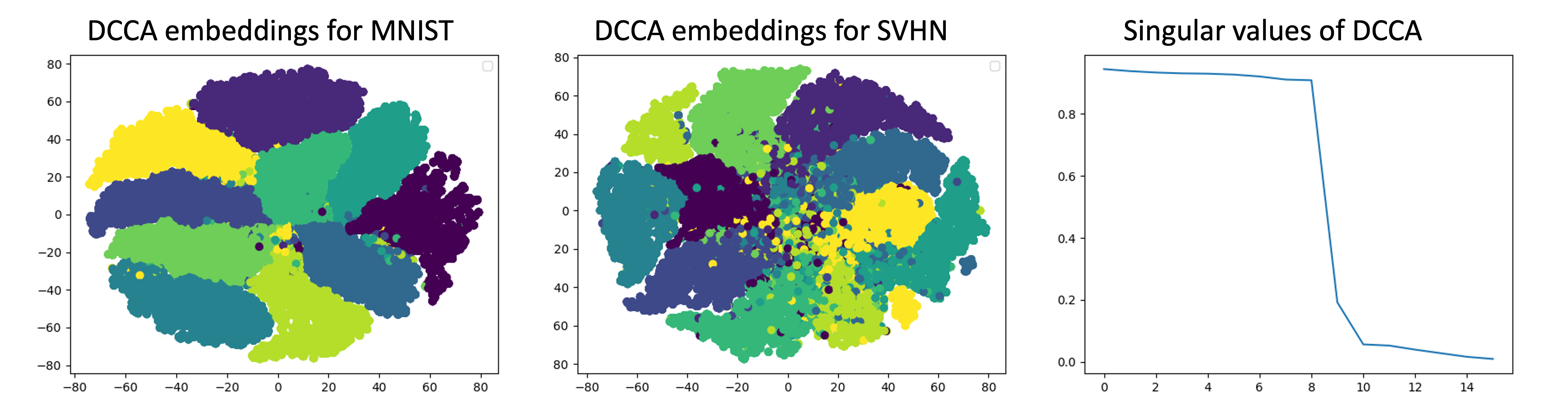}
    \caption{\small DCCA embeddings for the MNIST images $g_1(x_1)$ (on the left) and for SVHN images $g_2(x_2)$ (on the right). We use an output dimension of size 16 and project the embeddings in 2D with the TSNE \cite{van2008visualizing} algorithm. Each point represents an image and each color a label. On the right, we plot the singular values of $T \coloneqq \Sigma_1^{\frac{1}{2}}\Sigma_{1,2} \Sigma_{2}^{\frac{1}{2}}$ from the highest to the lowest. Those values represent the correlation that is contained in each direction after the DCCA is applied.}
\label{DCCA embeddings}
\end{figure}

In this section, we give additional results regarding the DCCA embeddings trained on MNIST-SVHN. Figure~\ref{DCCA embeddings} presents a 2D visualization of the embeddings learned for each modality. We see that the embeddings separate well the images according to their labels which is a desired property. We also present on the right the singular values of the 
matrix $T \coloneqq \Sigma_1^{\frac{1}{2}}\Sigma_{1,2} \Sigma_{2}^{\frac{1}{2}}$ the sum of which is optimized during the DCCA training. Those values represent the correlation that is contained in each direction of the embeddings. In this work, we use this plot to select the most correlated dimensions that we use in our embeddings. For instance, for the MNIST-SVHN dataset we choose to keep the 9 most correlated dimensions based on Figure~\ref{DCCA embeddings}. However, we also investigate how this choice impacts the performance of the JNF-DCCA model. 

\begin{figure}[ht]
    \centering
    \begin{minipage}[t]{.5\textwidth}
    \vspace{0.pt}
        \centering
        \includegraphics[width=\linewidth]{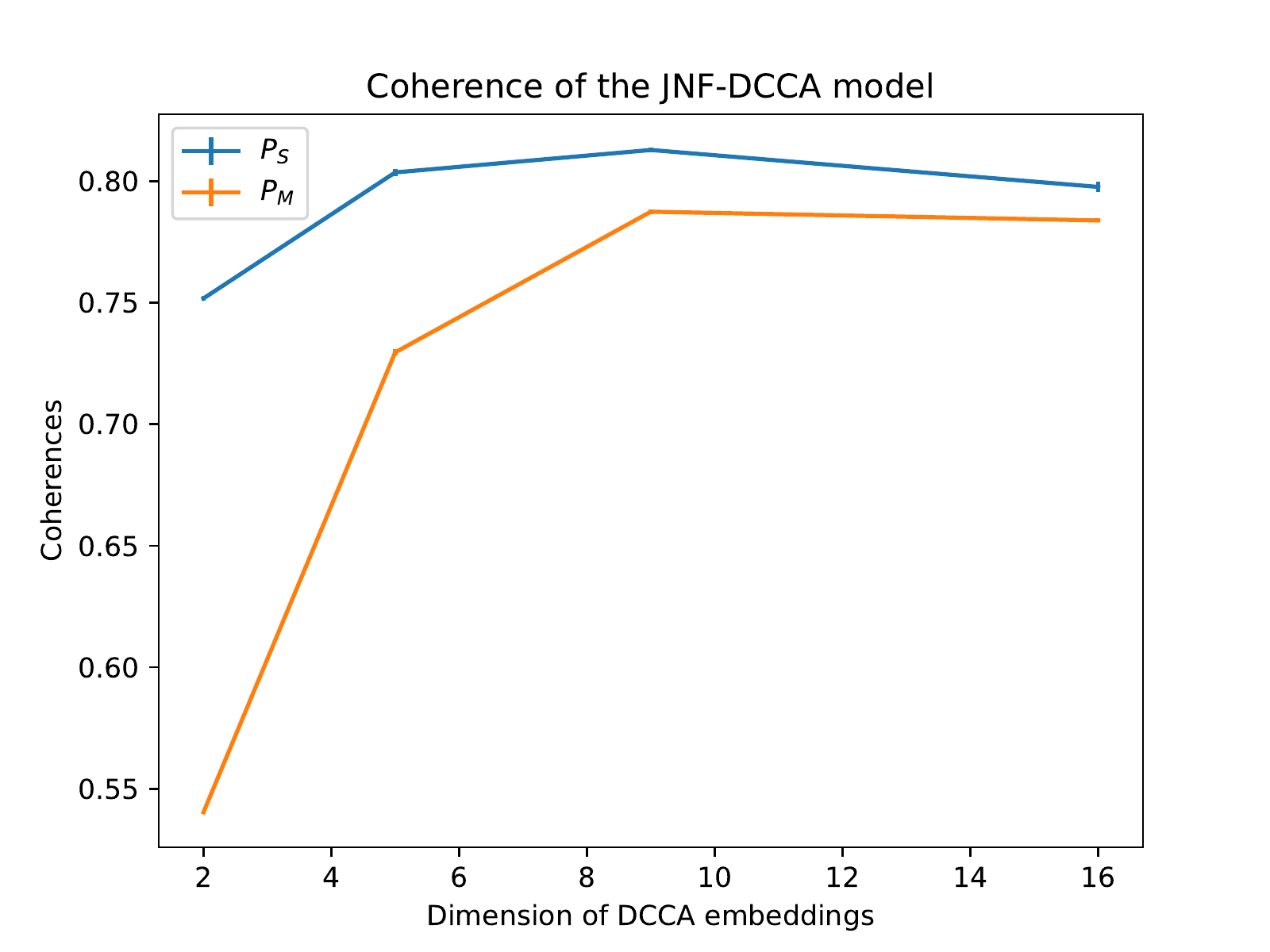}
        
    \end{minipage}%
    \begin{minipage}[t]{0.5\textwidth}
    \vspace{0pt}
        \centering
        \includegraphics[width=\linewidth]{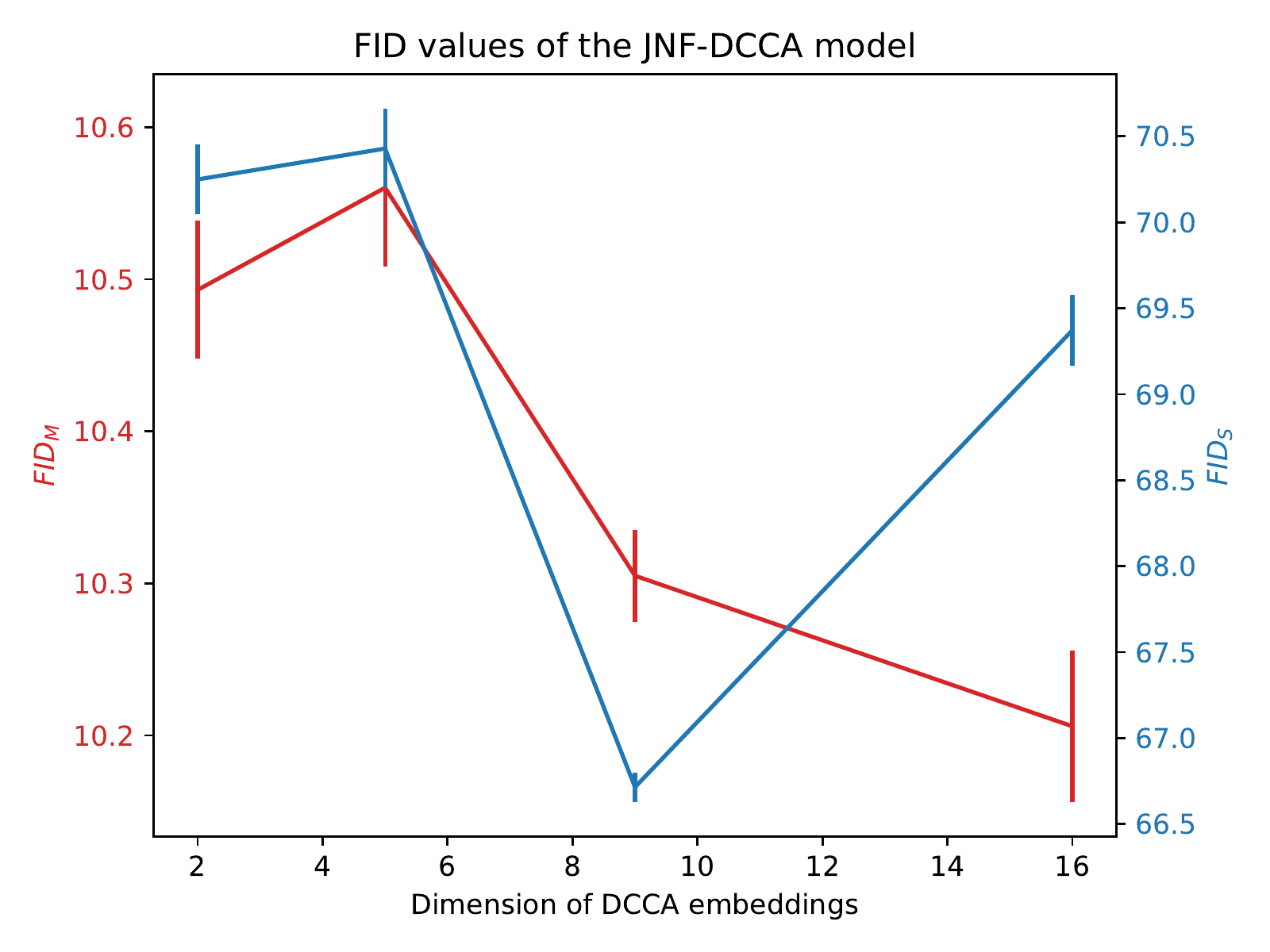}
        
    \end{minipage}
    \caption{Impact of the dimension of the DCCA embeddings on the performance of the JNF-DCCA model on the MNIST-SVHN dataset. \textit{Left}: Coherences as a function of the DCCA dimension. \textit{Right}: $\mathrm{FID}$ values as a function of DCCA dimension.}
    \label{impact of dcca dim}
\end{figure}

Figure.~\ref{impact of dcca dim} presents the coherences and $\mathrm{FID}$ results of the JNF-DCCA models depending on how many dimensions we keep in the DCCA embeddings. In each case, the most correlated dimensions are kept. We see that the optimal dimension is the one chosen based on Figure~\ref{DCCA embeddings} ($dim=9$). Taking a smaller dimension means loosing part of the shared information which results in a loss in coherence. Taking more dimensions correspond to adding noise and slightly diminishes the coherences. 

\clearpage

\section{Hamiltonian Monte Carlo Sampling}
\label{app:hmc}
In this appendix, we recall the principles of Hamiltonian Monte Carlo Sampling and detail how we apply it in our model. 
The Hamiltonian Monte Carlo (HMC) sampling belongs to the larger class of Markov Chain Monte Carlo methods (MCMC) that allow to sample from any distribution $f(z)$ known up to a constant. 
The general principle is to build a Markov Chain that will have $f(z)$ as stationary distribution. More specifically, the HMC is an instance of the Metropolis-Hasting Algorithm (see \ref{alg:metropolis}) that uses a physics-oriented proposal distribution. 
\begin{algorithm}[htb]
   \caption{Metropolis-Hasting Algorithm}
   \label{alg:metropolis}
\begin{algorithmic}[1]
   \STATE {\bfseries Initialization : $z \gets z_0$}
   \FOR{$i\coloneqq0$ $\rightarrow$ N} 
   \STATE {Sample $z'$ from the proposal $g(z'|z)$}
   \STATE {With probability $\alpha(z',z)$ accept the proposal $z\gets z'$}
   \ENDFOR
\end{algorithmic}
\end{algorithm}

Sampling from the proposal distribution $g(z'|z_0)$ is done by integrating the Hamiltonian equations : 
\begin{equation}
    \label{HMC-evol}
    \left\{ \begin{aligned}
        & \frac{\partial z}{\partial t} = \frac{\partial H}{\partial v}\,,\\
        & \frac{\partial v}{\partial t} = -\frac{\partial H}{\partial z}\,,\\
        & z(0) = z_0\,\\
        & v(0) = v_0 \sim \mathcal{N}(0,I)\,,\\
    \end{aligned}\right.
\end{equation}
where the Hamiltonian is defined by $H(z,v) = - \ln f(z) + \frac{1}{2}v^tv$. 
In physics, Eq.~\eqref{HMC-evol} describes the evolution in time of a physical particle with initial position $z$ and a random initial momentum $v$. The leap-frog numerical scheme is used to integrate Eq.~\eqref{HMC-evol} and is repeated $l$ times with a small integrator step size $\epsilon$ :
\begin{equation}
    \label{leapfrog}
    \begin{split}
        &v(t + \frac{\epsilon}{2}) = v(t) + \frac{\epsilon}{2} \cdot \nabla_z(\ln f(z)(t))\,,\\
        &z(t+\epsilon) = z(t) + \epsilon \cdot v(t +\frac{\epsilon}{2})\,,\\
        &v(t+ \epsilon) = v(t+\frac{\epsilon}{2}) + \frac{\epsilon}{2} \nabla_z \ln f(z(t+\epsilon))\,.\\
    \end{split}
\end{equation}
After $l$ integration steps, we obtain the proposal position $z'= z(t + l\cdot\epsilon)$ that corresponds to step $3$ in Algorithm \ref{alg:metropolis}. The acceptance ratio is then defined as $\alpha(z',z_0) = \min\left(1, \frac{\exp(-H(z_0,v_0))}{\exp(-H(z',v(t + l\cdot\epsilon)))}\right)$. 
This procedure is repeated to produce an ergodic Markov chain $(z^n)$ converging to the target distribution $f$ \cite{duane_hybrid_1987, liu_monte_2008, neal_mcmc_2011, girolami_riemann_2011}. In this work, we use HMC sampling to sample from the PoE of unimodal posteriors in Eq.~\eqref{approw poe}. To do so we need to compute and derivate the (log) of the target distribution given by the PoE of the unimodal distributions:
\begin{equation}
\ln q(z|(x_i)_{i\in S}) = -\ln p(z) + \sum_{i \in S} \ln q_{\phi_i}(z|x_i)\,.
\end{equation}
We can use autograd to automatically compute the gradient $\nabla_z \ln q(z|(x_i)_{i\in S})$ that is needed in the leapfrog steps.

\clearpage

\section{Experiments Details and Architectures}

\begin{figure}[ht]
    \centering
    \includegraphics[width = \linewidth]{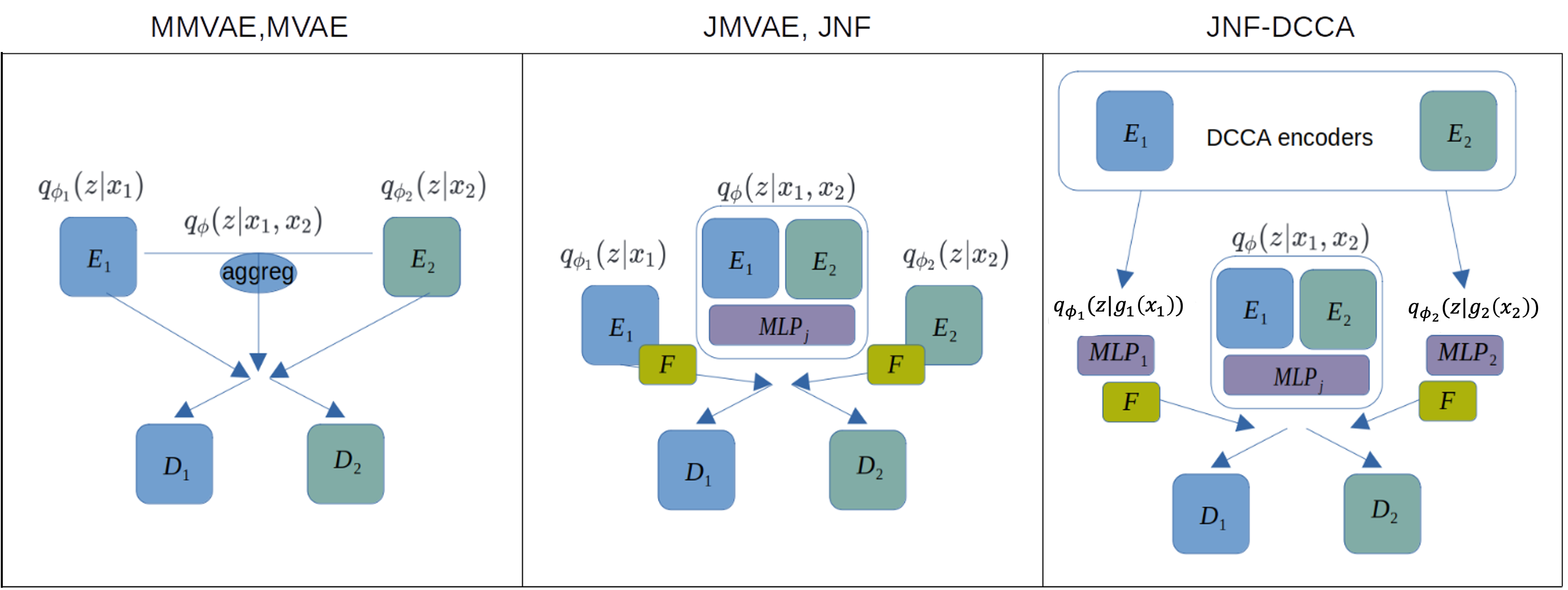}
    \caption{Summary of the components in each method. Each rectangle represent a network : $E$ for encoders, $F$ for the flows and $D$ for decoders. In the JMVAE, the $F$ block is omitted. Blocks with the same name have the same architecture. The precise architectures of the blocks depend on the dataset and are specified in each subsection of the appendices.}
    \label{fig:archi_components}
\end{figure}

In this appendix, we provide additional details on our experimental set-up. We summarize the architectures components of each model in Figure~\ref{fig:archi_components}. The networks architectures are described in the subsections dedicated to each dataset. For the MMVAE model, we use the original implementation available on github (\url{https://github.com/iffsid/mmvae}). We use our own implementations of the JMVAE, MVAE model. 
Our code is partly based on the Pytorch DCCA implementation available at \url{https://github.com/Michaelvll/DeepCCA} and also uses part of the code from this repository (\url{https://github.com/mseitzer/pytorch-fid}) to compute $\mathrm{FID}$ scores. 
Our models JNF and JNF-DCCA are implemented based on the Pytorch Library \cite{paszke2017automatic} and the Pythae Library \cite{chadebecpythae}.

The models are trained with either an 32GB V100 GPU or a 15GB RTX6000 GPU. We use WandB \cite{wandb} to monitor the trainings of the models.

\label{architectures}

\subsection{MNIST-SVHN}

For the MNIST-SVHN dataset, we pair each image with 5 different images with the same label in the other modality so that the total training set contains 270340 samples, the testing set 50000 images and the validation set 10000 images.

For $i=1,2$ the decoder distribution $p_\theta(x_i|z)$ is modelled as a Normal distribution $\mathcal{N}(\mu_i(z),I)$ where $\mu_i(z)$ is the output of the decoder network $D_i$ which architecture is detailed in Table~\ref{archi ms}. In the JMVAE, MMVAE, MVAE models, the encoder distributions are modelled as Normal distributions $\mathcal{N}(\nu_i(x_i), \sigma_i(x_i))$ with $\sigma_i(x_i)$ being a diagonal matrix. $\nu_i(x_i)$ and $\sigma_i(x_i)$ are the outputs of the encoders $E_i$ which architectures are given in Table~\ref{archi ms}.

 For JMVAE, JNF, and JNF-DCCA, we use a two-steps training with 100 epochs for the first step. All models are trained for a total of 200 epochs with an initial learning rate of $1\times10^{-3}$ and batchsize 128. The DCCA encoders have their dedicated training lasting 100 epochs, with batchsize 800, learning rate $1\times10^{-3}$ and embedding size 9. The influence of the embedding size on the performance is further studied in Appendix~\ref{app:dcca_embeddings}. For all models, the reconstruction terms for each modality are rescaled to give more weights to the MNIST-images: the reconstruction term $\ln p_\theta(x_1|z)$ is multiplied by  $\frac{3\times32\times32}{1\times28\times28}$. This rescaling follows the implementation of \cite{wu2018multimodal, shi2019variational} for the MMVAE and MVAE and has shown beneficial to the training of our models as well. 
 The architectures of each model is summarized in the Figure~\ref{fig:archi_components}. The specification of each block for the MNIST-SVHN experiments is detailed in Table~\ref{archi ms}.

\begin{table}[t]
\vskip 0.15in
\begin{center}
\begin{small}
\begin{sc}
\begin{tabular}{lcccr}
\toprule
$E_1$ & $E_2$ & $D_1$ & $D_2$ \\
\midrule
 Linear(784,512) + ReLu  & Conv(32@4x4,2,1)+ReLu& Linear(20,512)+ReLu& DConv(128,1,0)+Relu \\
  Linear(512,$\mathrm{dim}_1$)  &  Conv(64@4x4,2,1)+ReLu  & Linear(512,784) +Sigmoid&DConv(64,2,1)+Relu  \\
     & Conv(128@4x4,2,1)+ReLu &    & Dconv(32,2,1)+Relu\\
       &  Conv($\mathrm{dim}_2$ @ 4x4)  &     & Dconv(3,2,1)+Sigmoïd\\
\midrule
$MLP_j$ & $MLP_1$ & $MLP_2$ & $F$ \\
\midrule
 Linear(40,512)+ReLu&Linear($d_{\mathrm{DCCA}}$,512)+ReLu& Linear($d_{\mathrm{DCCA}}$,512)+ReLu& MAF flows : \\
  Linear(512,20)&Linear(512,512)+ReLu  & Linear(512,512)+ReLu  &  2 MADE blocks(3,128)\\
     & Linear(512,512)+ReLu& Linear(512,512)+ReLu& \\
       &  Linear(512,20)& Linear(512,20)&\\
\bottomrule
\end{tabular}
\end{sc}
\end{small}
\end{center}
\vskip -0.1in
\caption{Neural architectures of all the blocks used in the MNIST-SVHN experiment. This table is to be read with Figure \ref{fig:archi_components}. $\mathrm{dim}_1, \mathrm{dim}_2=d_{\mathrm{DCCA}}$ when $E_1$ (resp $E_2$) is used in the DCCA encoder otherwise $\mathrm{dim}_1, \mathrm{dim}_2=20$. In the results presented in the paper $d_{\mathrm{DCCA}}=9$. In Appendix~\ref{app:dcca_embeddings}, we vary this dimension. When the encoder $E_1$, $E_2$ are used to parameterize Normal distributions, the last layer is doubled to output the mean and the log-covariance.  }
\label{archi ms}
\end{table}

\subsection{CelebA}

For JMVAE, JNF, and JNF-DCCA, we use a two-steps training with 30 epochs for the first step. All models are trained for a total of 60 epochs with an initial learning rate of $1\times10^{-3}$ and batchsize 256. The latent space is chosen of dimension 64 as in \cite{suzuki2016joint}. The DCCA encoders have their dedicated training lasting 100 epochs, with batchsize 800, learning rate $1\times10^{-3}$ and embedding size of 40. 
For the MMVAE and MVAE models, the reconstruction terms for each modality are rescaled to balance the weight of each modality since they are of different sizes \emph{i.e}: the reconstruction term $\ln p_\theta(x_2|z)$ is multiplied by 50 for the MVAE and by $\frac{3\times64\times64}{40}$ for MMVAE. This rescaling is necessary to avoid modality collapse for those models. This phenomenon is explained by conflictual gradients in \cite{javaloy2022mitigating}. 

For the CelebA images, the decoder distribution $p_\theta(x_1|z)$ is modelled as a Normal distribution $\mathcal{N}(\mu_1(z),I)$ where $\mu_1(z)$ is the output of the decoder network $D_1$ which architecture is detailed in Table~\ref{archi ca}. For the binary vectors of attributes, the decoder distribution $p_\theta(x_2|z)$ is modelled as a Bernoulli distribution with parameters $p_2(z)$ that is the output of the decoder network $D_2$ (see Table~\ref{archi ca}).
In the JMVAE, MMVAE, MVAE models, the encoder distributions are modelled as Normal distributions $\mathcal{N}(\nu_i(x_i), \sigma_i(x_i))$ with $\sigma_i(x_i)$ being a diagonal matrix. $\nu_i(x_i)$ and $\sigma_i(x_i)$ are the outputs of the encoders $E_i$ which architectures are given in Table~\ref{archi ca}.

The architectures of each model is summarized in the Figure.~\ref{fig:archi_components}. The specification of each block for the CelebA experiments are detailed in Table.~\ref{archi ca}.

\begin{table}[t]
\caption{Neural architectures of all the blocks used in the CelebA experiment. This table is to be read with Figure.~\ref{fig:archi_components}. $\mathrm{dim}_1 = 40$ when $E_1$ is used in the DCCA encoder, $\mathrm{dim}_1 = 128$ when $E_1$ is used in the joint encoder, otherwise $\mathrm{dim}_1 = 64$. $\mathrm{dim}_2 = 40$ when $E_2$ is used in the DCCA encoder or in the joint encoder, otherwise $\mathrm{dim}_2=64$. When the encoder $E_1$, $E_2$ are used to parametrize normal distribution, the last layer is doubled to output the mean and the log-covariance.}
\label{archi ca}
\vskip 0.15in
\begin{center}
\begin{small}
\begin{sc}
\begin{tabular}{lcccr}
\toprule
$E_1$ & $E_2$ & $D_1$ & $D_2$ \\
\midrule
  Conv(64,4,2)&  Linear(40,512)+ReLu  & Linear(64,2048)   & Linear(64,512)+ReLu\\
   Conv(128,4,2)& Linear(512,$\mathrm{dim}_2$)  & ConvT(128,3,2)  & Linear(512,40)+Sigmoid\\
    Conv(128,3,2) &  & ResBlock**   & \\
     Conv(128,3,2)  &    & ResBlock**     & \\
     ResBlock**& & ConvT(128,5,2)+Sigmoid &\\
     ResBlock**& & ConvT(64,5,2)+Sigmoid&\\
     Linear(2048,$\mathrm{dim}_1$)& & ConvT(3,4,2) + Sigmoid&\\
\midrule
$MLP_j$ & $MLP_1$ & $MLP_2$ & $F$ \\
\midrule
 Linear(168,512)+ReLu  & Linear($d_{\mathrm{DCCA}}$, 512)+ReLu& Linear($d_{\mathrm{DCCA}}$,512)+ReLu   & MAF flows : \\
  Linear(64)  &  Linear(512,512)+ReLu  & Linear(512,512)+ReLu  &  2 MADE blocks(3,128)  \\
     & Linear(512,512)+ReLu & Linear(512,512)+ReLu   & \\
       &  Linear(512,64)  & Linear(512,64)    &\\

\bottomrule
\end{tabular}
\end{sc}
\end{small}
\end{center}
\vskip -0.1in
\end{table}

\subsection{MNIST-SVHN-FashionMNIST}

For the MNIST-SVHN-FashionMNIST dataset we do 5 differents random matchings between modalities so that the training dataset contains 275975 samples (from which we set 10000 images aside for validation) and the test dataset contains 48930 images. 

For $i=1,2,3$ the decoder distribution $p_\theta(x_i|z)$ is modelled as a Normal distribution $\mathcal{N}(\mu_i(z),I)$ where $\mu_i(z)$ is the output of the decoder network $D_i$ which architecture is detailed in Table~\ref{archi ms}. In the JMVAE, MMVAE, MVAE models, the encoder distributions are modelled as Normal distributions $\mathcal{N}(\nu_i(x_i), \sigma_i(x_i))$ with $\sigma_i(x_i)$ being a diagonal matrix. $\nu_i(x_i)$ and $\sigma_i(x_i)$ are the outputs of the encoders $E_i$ which architectures are given in Table~\ref{archi ms} where we consider that the same encoder and decoder blocks are used for MNIST and for FashionMNIST. 
For the MVAE training we use the subsampling paradigm used in the original article. \cite{wu_multimodal_2018}.

For JMVAE, JNF, and JNF-DCCA, we use a two-steps training with 100 epochs for the first step. All models are trained for a total of 200 epochs with an initial learning rate of $1\times10^{-3}$ and batchsize 128. The DCCA encoders have their dedicated training lasting 100 epochs, with batchsize 800 and learning rate $1\times10^{-3}$. For the MMVAE and MVAE models, the reconstruction terms for each modality are rescaled to balance the weight of each modality since they are of different size \emph{i.e}: the reconstruction terms for $\ln p_\theta(x_1|z)$ and $\ln p_\theta(x_3|z)$ is multiplied by  $\frac{3\times32\times32}{1\times28\times28}$.  
 The architectures of each model are summarized in the Figure~\ref{fig:archi_components}. The specification of each block for the MNIST-SVHN-FASHION experiments is detailed in Table~\ref{archi ms} where we consider that the same encoder and decoder blocks are used for MNIST and for FashionMNIST. 
    

\clearpage

\section{Training Paradigm for the JMVAE Model}
\label{training_compare}
In this appendix, we provide results on the influence of the training paradigm on the JMVAE model.
The JMVAE model uses warmup during training which forces the optimization of only the reconstruction term first to avoid local minima. The weight of the regularization is increased linearly to reach the value of 1 after $N_t$ epochs. Another particularity of the training is that all components are trained at the same time. On the other hand, we choose a two-steps training to train the joint encoder and decoders first and then fix those parameters after $N_t$ epochs when we start training the unimodal encoders. Here we compare the results obtained in each case:
\begin{itemize}
    \item Using the original one-step training with linear warmup,
    \item Using our two-steps training which dissociates the training of the joint encoder and decoders from the training of the unimodal encoders.
\end{itemize}
The first training paradigm requires choosing an hyperparameter $\alpha$ that controls a trade-off between reconstruction and cross-modal coherence. We choose the intermediate value $\alpha=0.1$ and also make the experiment with $\alpha=1$ to see the influence of the regularization. 

For the MNIST-SVHN dataset, we train the model for 200 epochs with $N_t=100$.
Table \ref{training_ms_compare} shows that the one-step training does not improve significantly the coherences compared to the two-steps training (as it improves $P_M$ but decreases $P_S$) but causes the $\mathrm{FID}s$ to significantly increase. This is because the $\mathcal{L}_{\mathrm{JM}}$ term regularizes two much the joint encoder $q_{\phi}(z|x_1,x_2)$ which impacts the quality of the reconstructions. This does not happen with the two steps training as $q_\phi(z|x_1,x_2)$ is already fixed when the $\mathcal{L}_{\mathrm{JM}}$ term is optimized.

\begin{table}[ht]
\caption{Coherence and $\mathrm{FID}$ values averaged on 5 runs. $P_M$ (resp $P_S$ is the coherence on MNIST (resp. SVHN) images generated from SVHN (resp. MNIST) images. The standard deviations are all $ \leq 0.003$ for the precision values and $\leq 0.5$ for the $\mathrm{FID}$ values. The $\mathrm{FID}$ are computed taking all the samples from the test dataset $\approx 50 000$}
\label{training_ms_compare}
\vskip 0.15in
\begin{center}
\begin{small}
\begin{sc}
\begin{tabular}{lccccccr}
\toprule
Model & $P_M$ & $P_S$ & $\mathrm{FID}_M$ & $\mathrm{FID}_S$& $P(S|M)$&$P(M|S)$&$P(S,M)$\\
\midrule
JMVAE $\alpha=0.1$ &\textbf{0.57} &0.71 & 18.8 & 89.5& $-741.2 \pm 0.3$& $-2848.0\pm0.7$&$-3594.5\pm 0.6$ \\
JMVAE $\alpha=1$ & 0.53 & 0.76 &   21.1& 83.8& $-740.8 \pm 0.3$ & $-2847.3 \pm 0.6$&$\textbf{-3590.1}\pm0.7$  \\
JMVAE-2steps    & 0.46 & \textbf{0.79} & \textbf{13.0} & \textbf{72.3}& \textbf{$\textbf{-740.5} \pm 0.6$}& \textbf{$-2847.1 \pm 1.3$}&$\textbf{-3590.5} \pm 1.3 $\\

\bottomrule
\end{tabular}
\end{sc}
\end{small}
\end{center}
\vskip -0.1in
\end{table}

The same phenomena happens on CelebA as shown in Table \ref{training_compare_c}. On this dataset all metrics are better for the two steps training. On this dataset we train for 60 epochs and choose $N_t=30$.

\begin{table}[ht]
\caption{Coherence and $\mathrm{FID}$ values averaged on 5 runs. The standard deviations are all $ \leq 0.003$ for the precision values and $\leq 0.5$ for the $\mathrm{FID}$ values. The $\mathrm{FID}$ are computed taking all the samples from the test dataset $\approx 20 000$ samples.}
\label{training_compare_c}
\vskip 0.15in
\begin{center}
\begin{small}
\begin{sc}
\begin{tabular}{lcccccr}
\toprule
Model & $P_C$ & $P_A$ & $\mathrm{FID}_C$ & $P(A|C)$ &$P(C|A)$&$P(A,C)$  \\
\midrule
JMVAE-$\alpha=0.1$&0.810& 0.861&100.7&$-10.9\pm0.2$ &$-11509 \pm4$ &$-11431 \pm2$\\     
JMVAE-$\alpha=1$ & 0.808 & 0.855 &  102.0&$-11.3\pm0.2$ &$-11500\pm3$& $-11427\pm2$\\
JMVAE- 2 steps    & \textbf{0.825} & \textbf{0.867} & \textbf{64.6}&$\textbf{-10.1}\pm0.2$&$\textbf{-11490}\pm 5$&$\textbf{-11414}\pm2$\\

\bottomrule
\end{tabular}
\end{sc}
\end{small}
\end{center}
\vskip -0.1in
\end{table}

\clearpage

\section{Ablation study : Influence of the number of flows}
\label{app:number of flows}
\begin{figure}[htb]
    \centering
    \begin{minipage}[t]{.45\textwidth}
    \vspace{0pt}

        \centering
        \includegraphics[width=\linewidth]{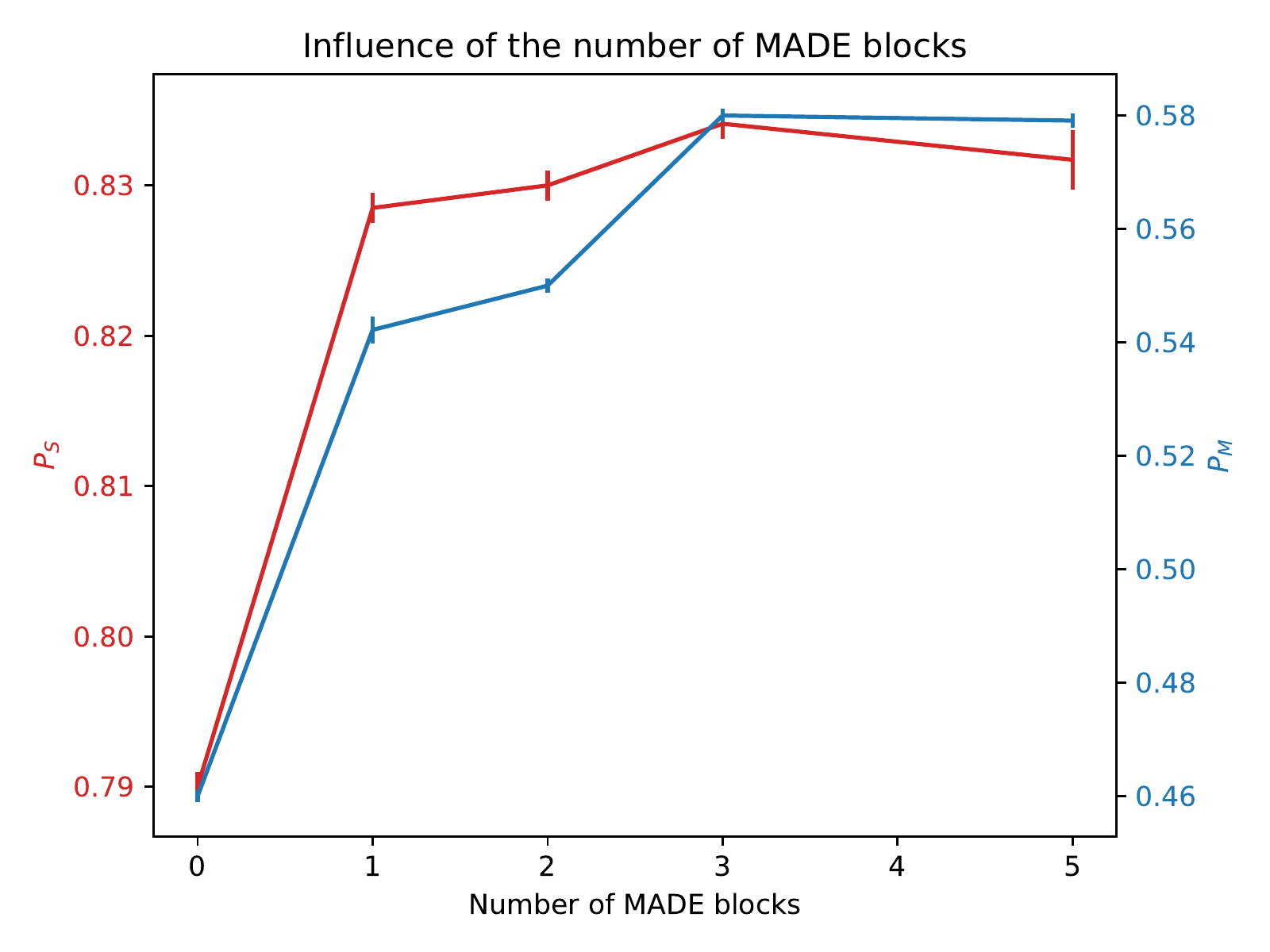}
    \end{minipage}
    \begin{minipage}[t]{0.45\textwidth}
    \vspace{0pt}

        \centering
        \includegraphics[width=\linewidth]{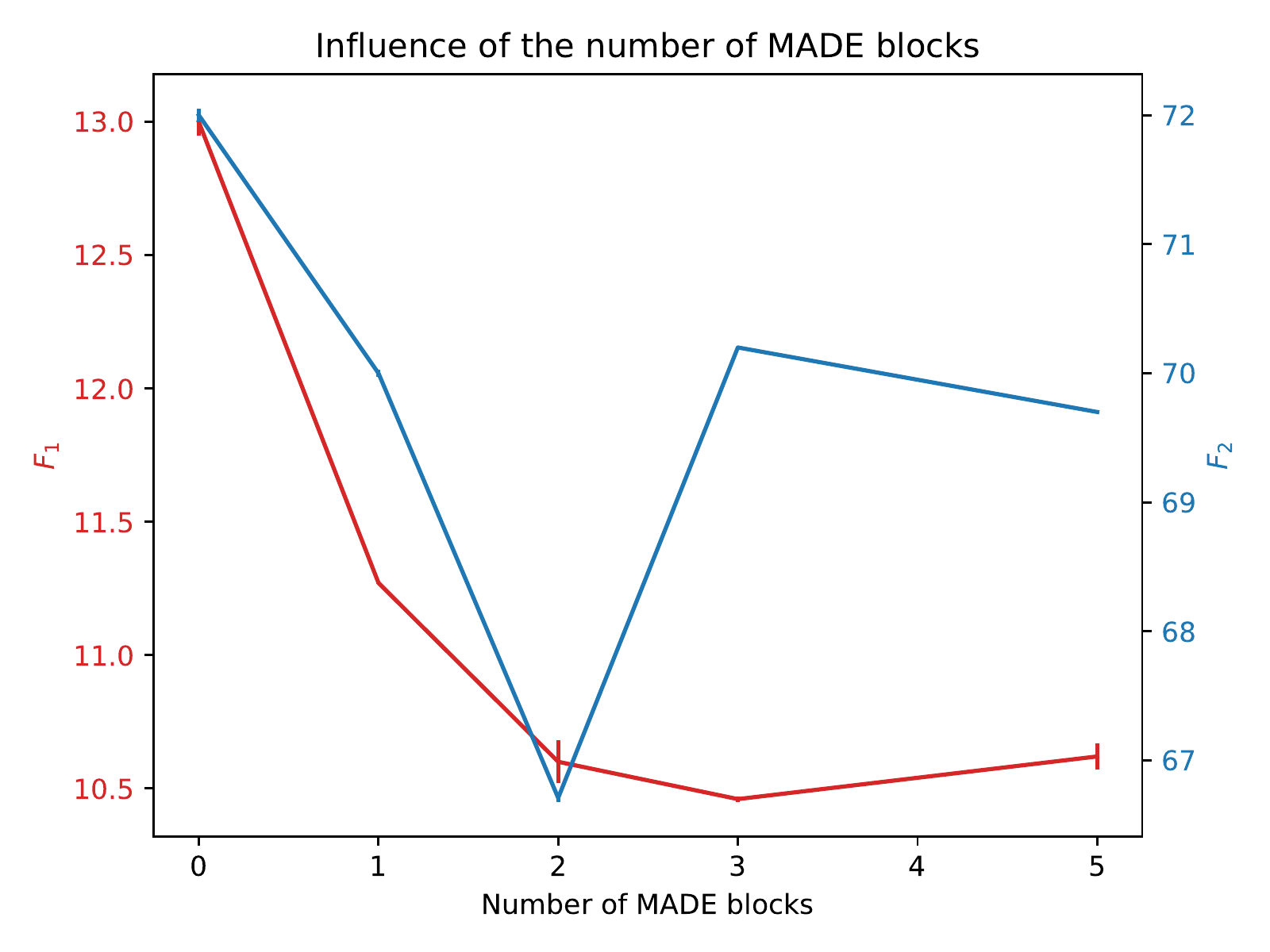}
    \end{minipage}
    \label{fig:nmade ablation on jnf}
    \caption{Influence of the number of MADE transformations in the MAF flows on the performance of the JNF model. \textit{Left}: Coherences of the model as a function of the number of MADE transformations. \textit{Right}: $\mathrm{FID}$ values as a function of the number of MADE transformations.}
\end{figure}

\begin{figure}[htb]
    \centering
    \begin{minipage}[t]{.45\textwidth}
    \vspace{0pt}

        \centering
        \includegraphics[width=\linewidth]{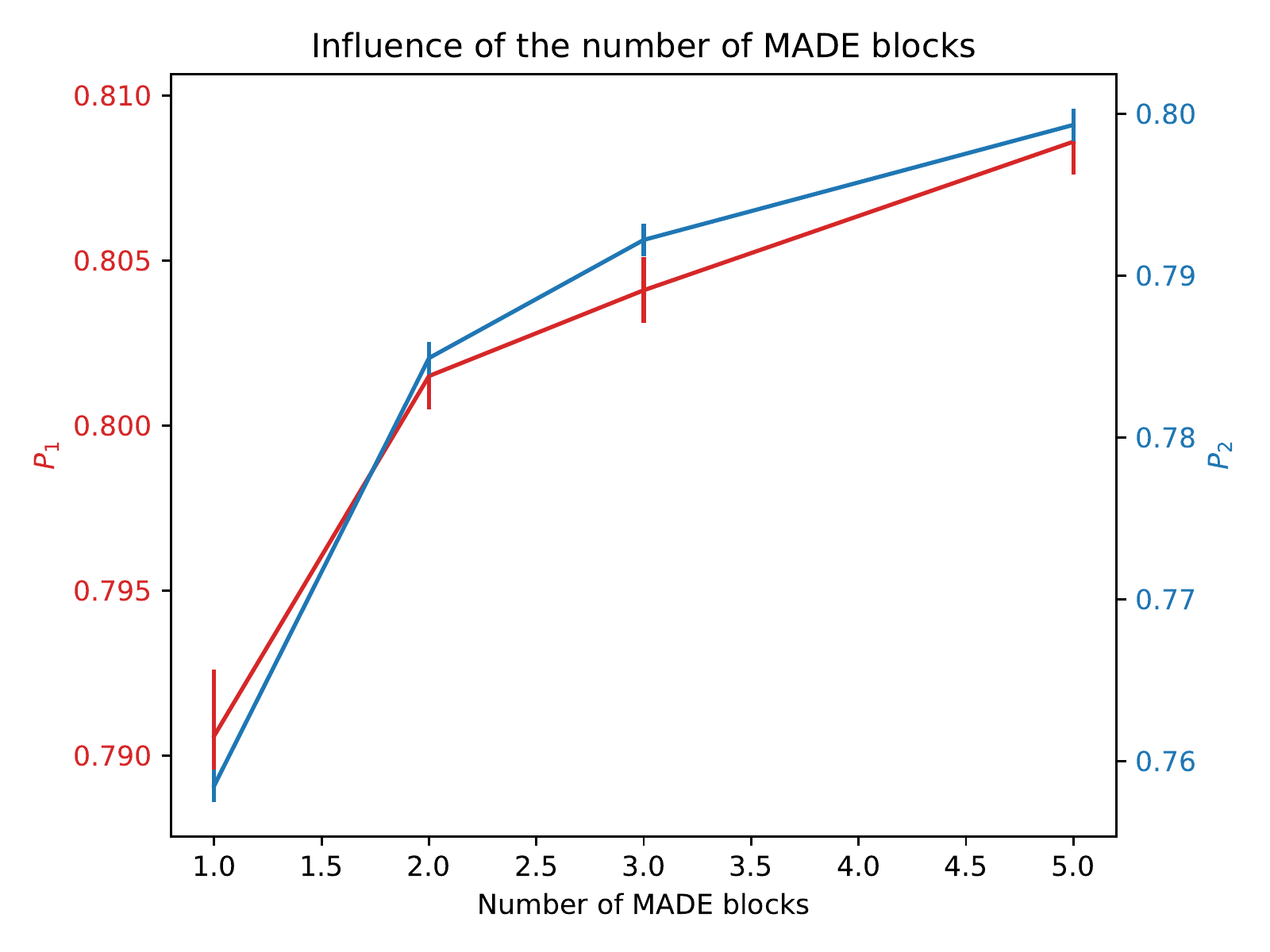}
    \end{minipage}
    \begin{minipage}[t]{0.45\textwidth}
    \vspace{0pt}

        \centering
        \includegraphics[width=\linewidth]{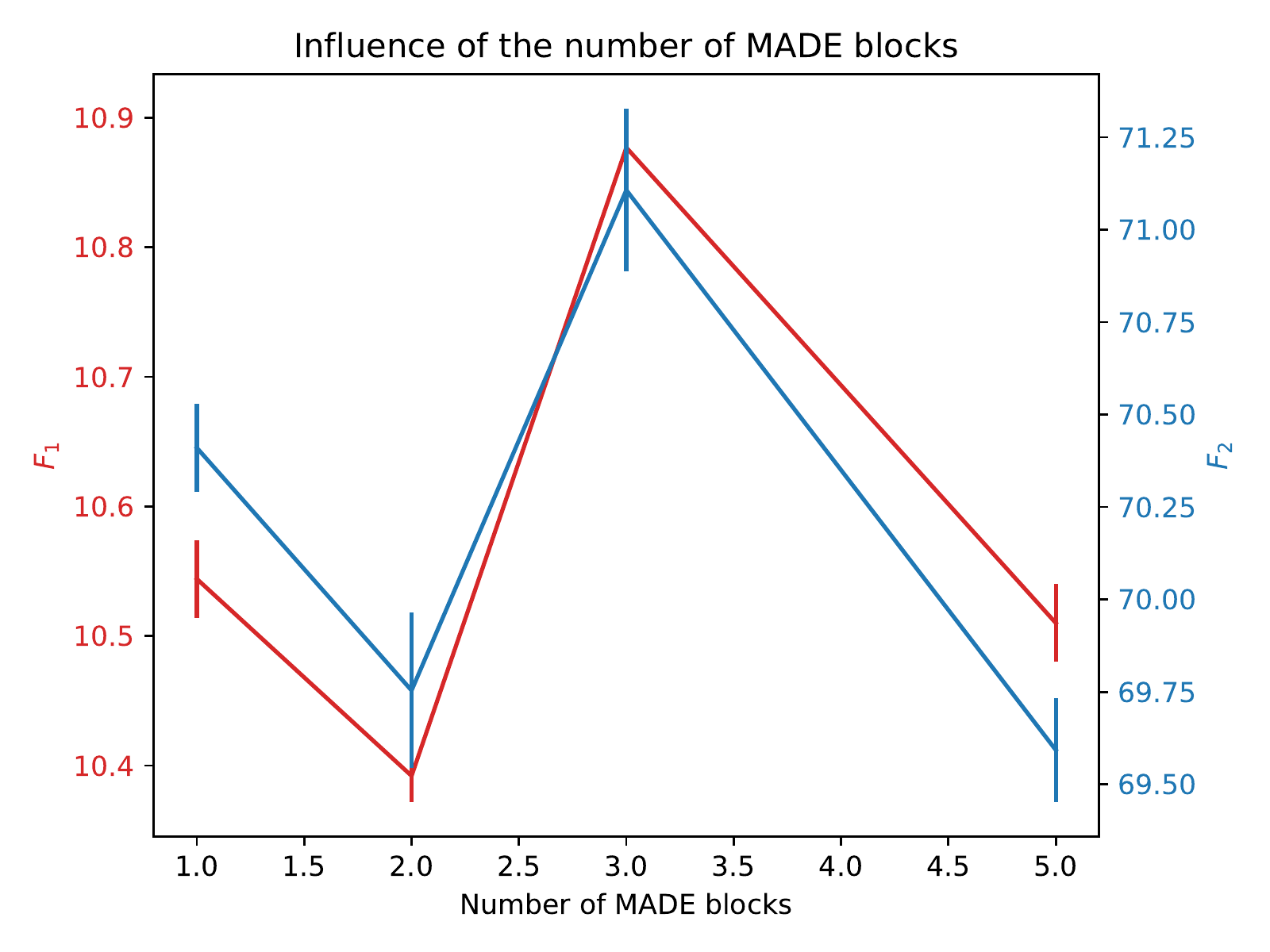}
    \end{minipage}
    \label{fig:nmade ablation on jnf-dcca}
    \caption{Influence of the number of MADE transformations in the MAF flows on the performance of the JNF-DCCA model. \textit{Left}: Coherences of the model as a function of the number of MADE transformations. \textit{Right}: $\mathrm{FID}$ values as a function of the number of MADE transformations.}
\end{figure}

In this appendix, we present the influence of the number of MADE blocks in the MAF flows on the performance of JNF, JNF-DCCA on the MNIST-SVHN dataset.
Figure~\ref{fig:nmade ablation on jnf} shows the results for the JNF model while Figure~\ref{fig:nmade ablation on jnf-dcca} displays the results for the JNF-DCCA model.

For both the JNF and JNF-DCCA models, it seems that augmenting the number of transformations in the flows improves the coherence.

\clearpage

\section{Classifiers Used for Coherence Computations}

In this appendix, we give details on the architectures and accuracies of the classifiers used in the evaluation of the models. For MNIST, SVHN and FashionMNIST we train from scratch the networks specified in Table~\ref{classifiers archi}.
For CelebA, we finetune the pre-trained network Resnet50 \cite{he2016deep}.

Table.~\ref{classifiers results} presents the accuracies of our classifiers on each dataset.    

\begin{table}[ht]
\caption{Neural architectures of all the classifiers}
\label{classifiers archi}
\vskip 0.15in
\begin{center}
\begin{small}
\begin{sc}
\begin{tabular}{lcr}
\toprule
MNIST/FashionMNIST & SVHN  \\
\midrule
  Conv(32,4,1)+BatchNorm+ReLu& Conv(32,4,1)+BatchNorm+ReLu\\ Conv(64,4,1)+BatchNorm+ReLu & Conv(64,4,1)+BatchNorm+ReLu   \\
   Linear(30976,512)+Dropout(0.5)&Conv(128,4,1)+BatchNorm+ReLu\\
       Linear(512,10)& Linear(67712,1024)+BatchNorm+Dropout(0.5)   \\
    & Linear(1024,512)+BatchNorm+Dropout(0.5)\\
     &Linear(512,10)\\

\bottomrule
\end{tabular}
\end{sc}
\end{small}
\end{center}
\vskip -0.1in
\end{table}

\begin{table}[ht]
\caption{Accuracies of the classifiers on the test datasets.}
\label{classifiers results}
\vskip 0.15in
\begin{center}
\begin{small}
\begin{sc}
\begin{tabular}{lccr}
\toprule
MNIST&FashionMNIST & SVHN & CelebA \\
\midrule
  0.98 & 0.95 &0.90 & 0.90 \\
   
\bottomrule
\end{tabular}
\end{sc}
\end{small}
\end{center}
\vskip -0.1in
\end{table}

\clearpage

\section{Computing Estimates for the Likelihoods}

The joint likelihood is estimated with the following Importance Sampling approximation with 1000 samples :
\begin{equation*}
    \begin{split}
    & p_{\theta}(X) = \int \prod_i p_{\theta}(x_i|z)p(z)dz\,,\\
    & \approx \frac{1}{n} \sum_{(z_k)_{k=1}^{n} \sim q_\phi(z| X)} \prod_i p_{\theta}(x_i|z_k)\frac{p(z_k)}{q_\phi(z_k |X)}\,.
    \end{split}.
\end{equation*}
The conditional likelihoods are estimated with the following Monte-Carlo approximation using 1000 samples : 
\begin{small}
    \begin{equation}
    p_{\theta,\phi_j}(x_i|x_j)  \approx \frac{1}{n}\sum_{(z_k)_{k=1}^n \sim q_{\phi_j}(z| x_j)} p_{\theta}(x_i|z_k)\,.
\end{equation}
\end{small}

\clearpage

\section{Additional Results on CelebA.}
\label{app:more celeba samples}
\begin{figure}[ht]
    \centering
    \includegraphics[width=0.7\textwidth]{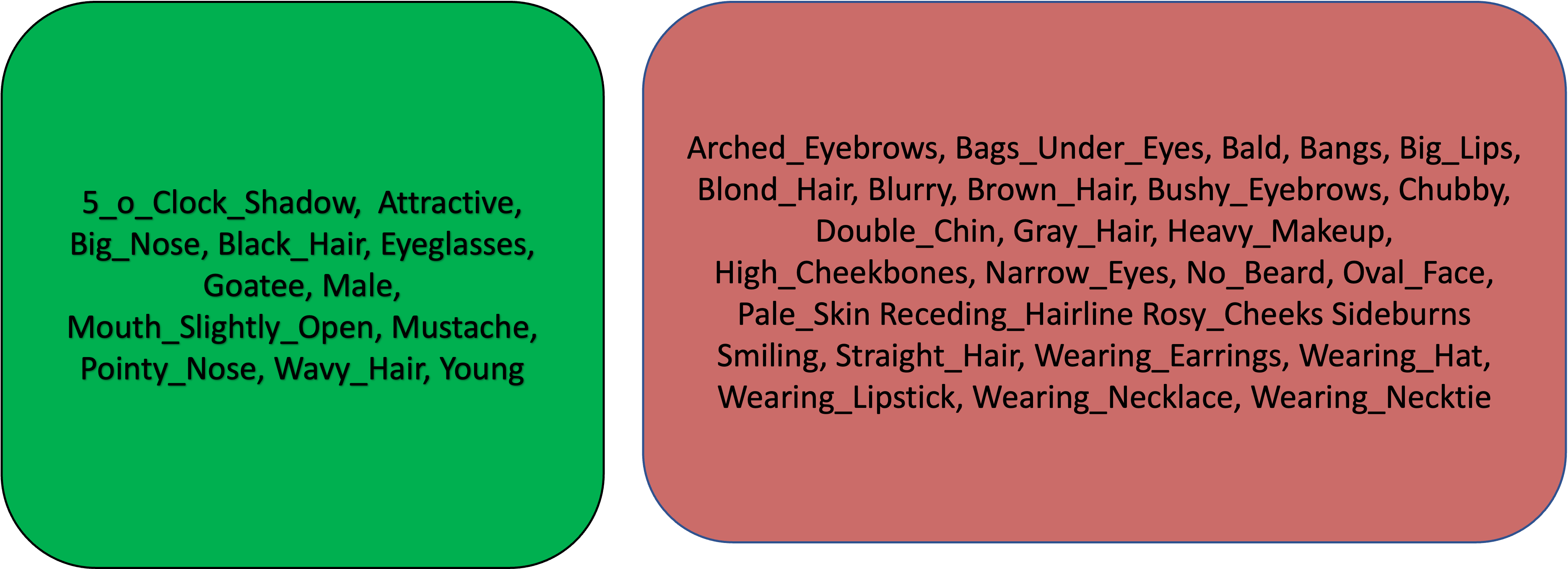}
    \caption{The attributes used to sample in Figure.~\ref{samples celeba}. The green attributes are present while the red attributes are absent.}
    \label{fig:attributes}
\end{figure}

\begin{figure}[ht]
    \centering
    \includegraphics[width=0.7\textwidth]{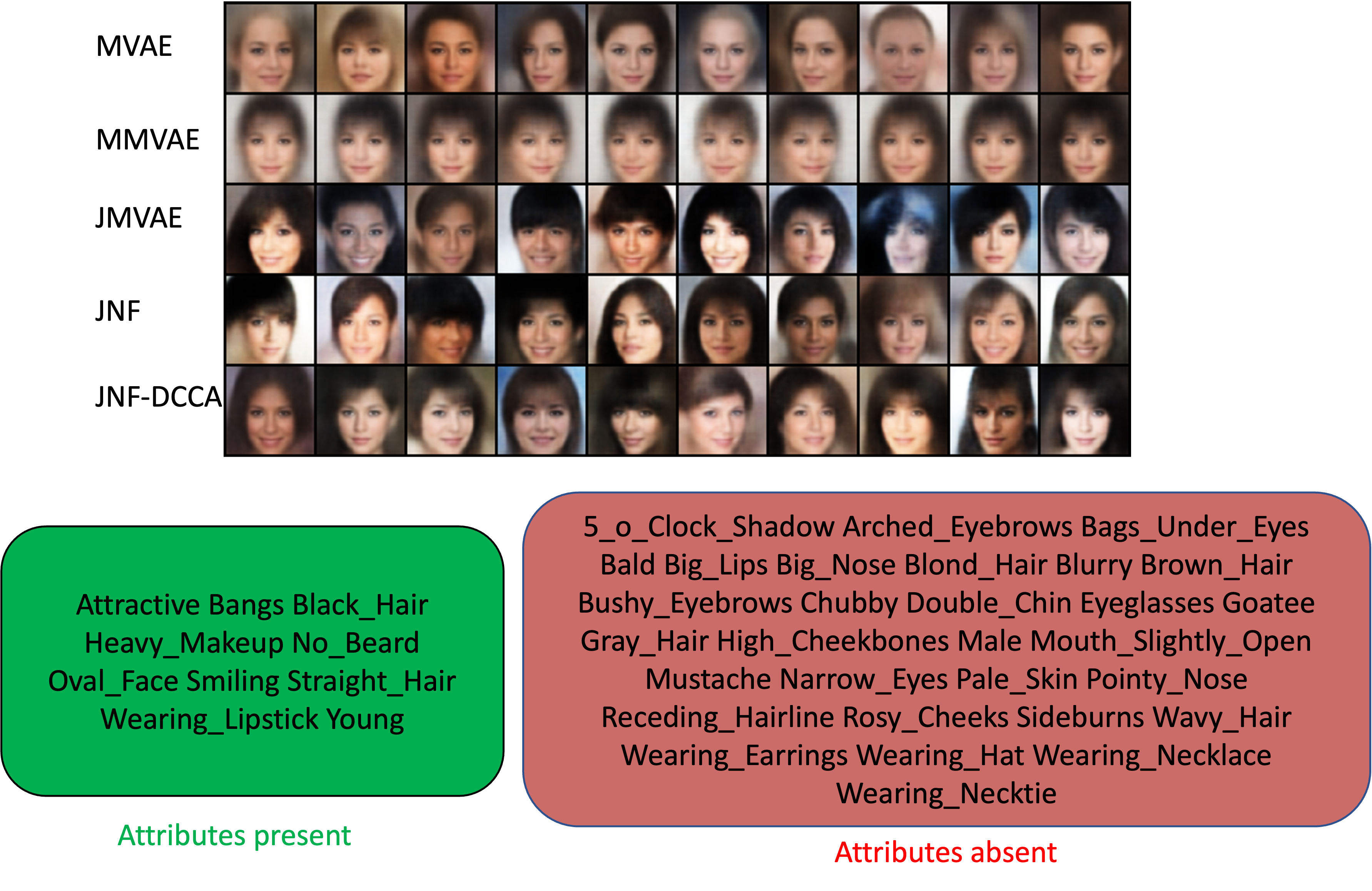}
    \caption{Another example of generating images from attributes.}
    \label{fig:another generation}
\end{figure}

\begin{figure}[ht]
    \centering
    \includegraphics[width=\textwidth]{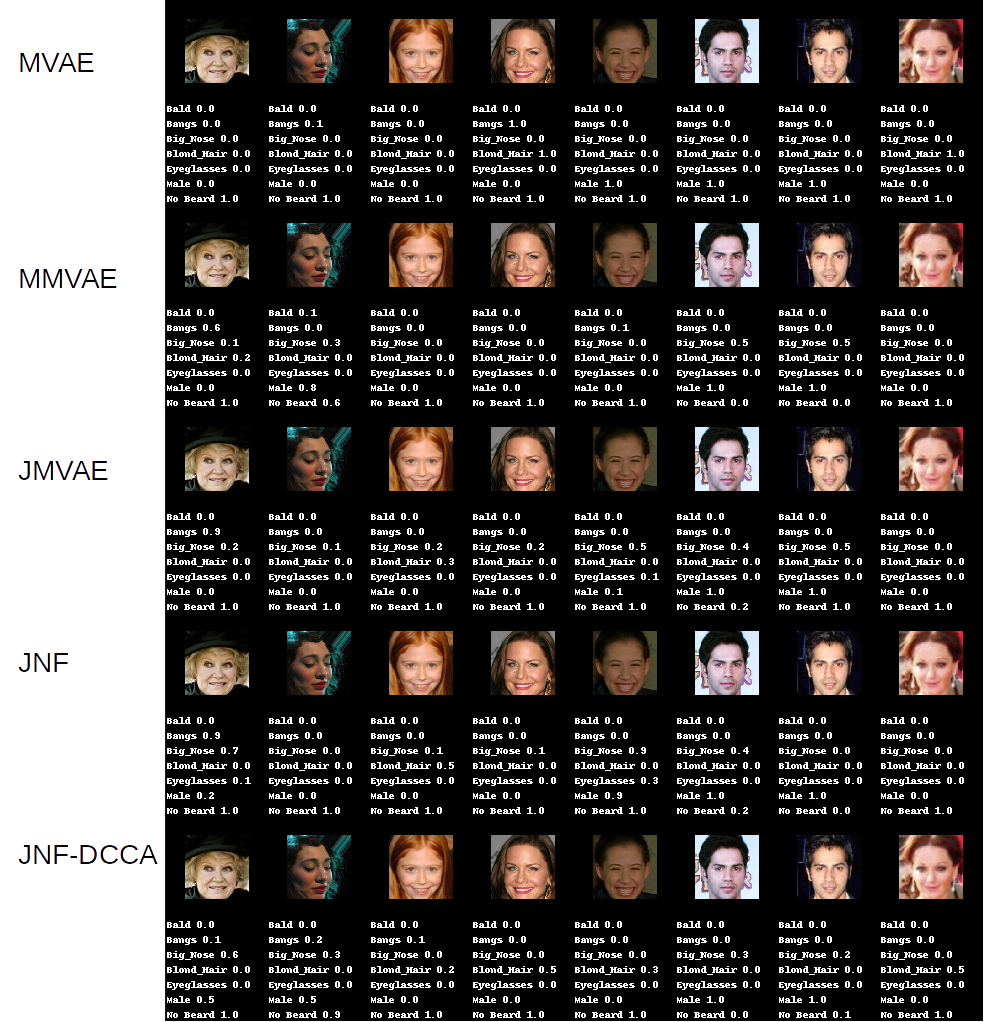}
    \caption{From the images, we generate a subset of attributes for each model. }
    \label{fig:generating attributes}
\end{figure}

In this section, we provide more experimental results for the CelebA experiment. First, we provide in Figure~\ref{fig:attributes} the entire set of attributes that is used to generate the samples in Figure~\ref{samples celeba}. We also provide another example of generating images from attributes in Figure.~\ref{fig:another generation}. Then we provide an example of generating attributes from images in Figure.~\ref{fig:generating attributes}. 

\subsection{Generating Attributes from Images}

\clearpage

\section{Additional Results on MNIST-SVHN-FashionMNIST}
 \label{add results on msf}

\begin{figure}
    \centering
    \includegraphics[width=\linewidth]{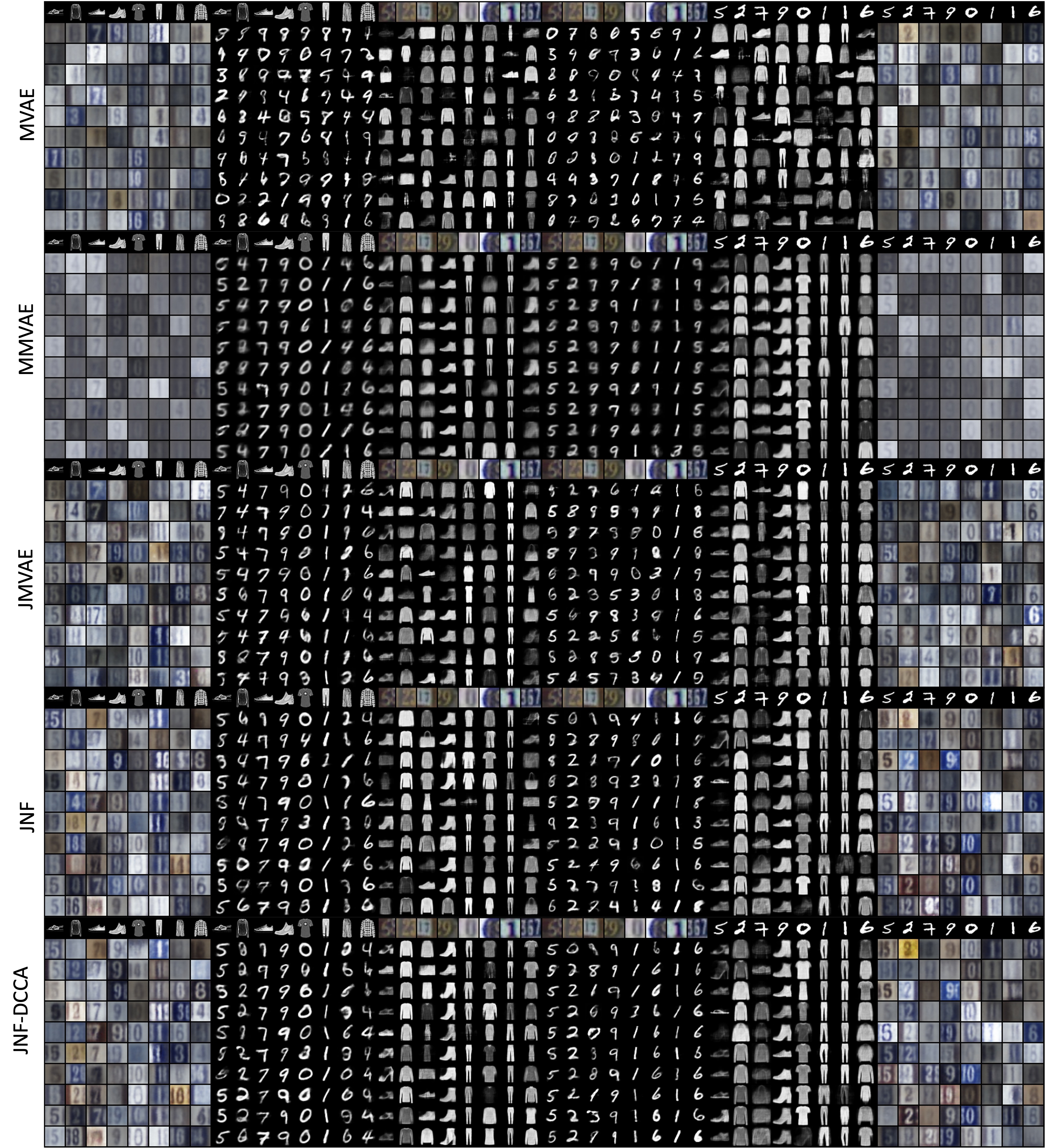}
    \caption{Conditional generations from one modality to another. For each model, the first line are the images we condition on and the following lines are generated samples conditioned on those images.}
    \label{fig:msf samples}
\end{figure}

In this section, we provide more qualitative and quantitative results on the trimodal dataset. Figure.~\ref{fig:msf samples} shows images generated from the conditional generations. We notice that with three modalities and even with rescaling, the MMVAE model is sensitive to modality collapse: the SVHN generations resemble averaged images and are not diversified. On the contrary, our models' generations are much more diversified. The JMVAE and MVAE models have a wider diversity than the MMVAE but the generations are less coherent, especially when it comes to generating MNIST images from another modality. 
The $\mathrm{FID}$ metrics presented in Table.~\ref{msf fids} confirms this qualitative analysis. 

\begin{table*}[ht]
\caption{The $\mathrm{FID}$ values computed using the entire test set composed of $\approx 50000$ samples.}
\label{msf fids}

\vskip 0.15in
\begin{center}
\begin{small}
\begin{sc}
\begin{tabular}{lcccccr}
\toprule
&S&F&M&F&M&S\\
Model &M&M&S&S&F&F\\
\midrule
MMVAE-30    & $62.0 \pm 0.1$ & $58.8 \pm0.1$ &  $212.2 \pm 0.4 $ & $202.3 \pm0.5$ &$105.8\pm0.2$& $110.1 \pm 0.1 $  \\
MVAE    & $\mathbf{16.7\boldsymbol{\pm}0.1}$& $\mathbf{16.9\boldsymbol{\pm}0.1}$& $93.1\pm0.2$ & $94.7\pm0.2$ & $66.1\pm0.1$ & $67.2\pm0.1$\\
JMVAE & $22.0\pm0.1$& $21.5\pm0.1$  &$\mathbf{59.0\boldsymbol{\pm} 0.2}$& $\mathbf{59.4\boldsymbol{\pm}0.1}$ & $\mathbf{65.3\boldsymbol{\pm}0.2}$& $69.1\pm0.1$ \\
JNF    & $22.2\pm0.1$& $20.6\pm0.1$ & $63.5\pm0.1$&$64.8\pm0.2$& $66.7\pm0.2$ & $69.1\pm0.2$ \\
JNF-DCCA     & $21.7\pm0.1$&$21.3\pm0.1$&$63.1\pm0.2$& $63.1\pm0.1$ &$67.8\pm0.1$ & $\mathbf{65.9\boldsymbol{\pm}0.2}$ \\

\bottomrule
\end{tabular}
\end{sc}
\end{small}
\end{center}
\vskip -0.1in
\end{table*}

\end{document}